\documentclass{midl} 

\usepackage{mwe} 
\usepackage{booktabs}
\usepackage{multicol}
\usepackage{floatrow}
\usepackage{indentfirst}
\usepackage{multirow}
\newfloatcommand{capbtabbox}{table}[][\FBwidth]
\jmlrvolume{-- Under Review}
\jmlryear{2021}
\jmlrworkshop{Full Paper -- MIDL 2021 submission}

\title[Deep CNNs for Peripheral Blood Cell Classification]{Deep CNNs for Peripheral Blood Cell Classification}

\midlauthor{\Name{Ekta Gavas\midljointauthortext{Equal Contribution}} \Email{ekta.gavas@research.iiit.ac.in}\\
\addr IIIT Hyderabad, India
\AND
\Name{Kaustubh Olpadkar\midlotherjointauthor} \Email{kolpadkar@cs.stonybrook.edu}\\
\addr Stony Brook University, NY, USA
}

\begin{document}

\maketitle

\begin{abstract}
The application of machine learning techniques to the medical domain is especially challenging due to the required level of precision and the incurrence of huge risks of minute errors. Employing these techniques to a more complex subdomain of hematological diagnosis seems quite promising, with automatic identification of blood cell types, which can help in detection of hematologic disorders. In this paper, we benchmark 27 popular deep convolutional neural network architectures on the microscopic peripheral blood cell images dataset. The dataset is publicly available, with large number of normal peripheral blood cells acquired using the CellaVision DM96 analyzer and identified by expert pathologists into eight different cell types. We fine-tune the state-of-the-art image classification models pre-trained on the ImageNet dataset for blood cell classification. We exploit data augmentation techniques during training to avoid overfitting and achieve generalization. An ensemble of the top performing models obtains significant improvements over past published works, achieving the state-of-the-art results with a classification accuracy of 99.51\%. Our work provides empirical baselines and benchmarks on standard deep-learning architectures for microscopic peripheral blood cell recognition task.
\end{abstract}


\section{Introduction}
Blood carries oxygen and nutrients to living cells in different organs and tissues. It carries away the waste for detoxification. It transports hormones to the desired site of action to fight infections and regulates body temperature. The ability to classify blood constituents can be critical in assessing the patient’s health. Plasma, which constitutes 55\% of blood, is a colored liquid comprising mainly water (about 90\%) and other essential substances such as proteins (albumin, clotting factors, antibodies, enzymes, and hormones), glucose, and fats. Rest 45\% of blood is composed of white blood cells (WBCs/leukocytes), platelets (thrombocytes), and red blood cells (RBCs/erythrocytes), which float in the plasma\cite{fathima17, deanl05}. All these cells are associated with different functionalities. RBCs are responsible for transporting gases ($O_{2}, CO_{2}$) from lungs to tissues and maintaining systemic acid/base equilibria. Damage of red cell integrity, defined as hemolysis, has been shown to significantly contribute to severe pathologies, including endothelial dysfunction\cite{kuhn17}.

Based on the presence of visible granules in the microscopic view, WBCs can be classified into two broad categories: granulocytes and agranulocytes (nongranulocytes). Neutrophils, eosinophils, and basophils belong to the granulocytes category, while lymphocytes and monocytes belong to the agranulocytes category \cite{Almezhghwi2020ImprovedCO,ACEVEDO2019105020}. Various types of WBCs play a role in immune response \cite{deanl05} and act as a defense mechanism in the body against illness-causing agents. Immature granulocytes (IG) are under-developed WBCs released from the bone marrow into the blood. Except for blood from newborn children or pregnant women, the appearance of IG (promyelocytes, myelocytes, and metamyelocytes) \cite{ACEVEDO2019105020} in the peripheral blood (PB) indicates an early-stage response to infection, inflammation, or other stimuli of the bone marrow. Similarly, erythroblasts are nucleated immature RBCs or erythroid precursors not seen after the neonatal period. Their appearance in PB of children and adults can signify bone marrow damage, stress, malignant neoplasms, or other potentially serious diseases \cite{constantino2000nucleated}. Platelets are anucleated cells in blood and get activated at the site of injury to form a blood clot. Besides, they play an important role in innate immunity and regulation of tumor growth and extravasations in the vessel\cite{holinstat2017normal}. They make up less than 1\% of blood volume.
Usually, the typical percentages of neutrophils in the blood are 0-6\%. Eosinophils constitute 1–3\%, basophils 0–1\%, lymphocytes 25–33\%, and monocytes 3–10\% of the leukocytes circulating in the blood\cite{ACEVEDO2019105020}.

Recognition of various blood cell types can reveal anomalous blood cell populations like immature cells (IG or erythroblasts). On accurate identification, differential blood cell count can suggest any possible abnormalities in the blood, or help diagnose an infection, inflammation, leukemia\cite{shafique2018computer, mathur2013scalable}, or any immune system disorder. Analyzing the blood cell morphology is the outset for the diagnosis of 80\% of hematological diseases\cite{ACEVEDO2019105020}. Quantitative morphological analysis can thus, help cytologists assess blood samples and conclude about the patient's blood conditions. However, the above processes are complex and time-consuming, involving a specialist meticulously examining the blood smear under a microscope, subjecting it to human errors. For the past few years, several attempts have been made to automate these processes using image processing techniques and machine learning, making them time and cost-effective and substantially reducing the workload in laboratories.

Convolutional neural networks (CNNs) are known to show excellent results on image recognition tasks, and hence, there has been an extensive research for applying them in the medical domain. In this paper, we explore various deep CNNs for blood cell classification task with peripheral blood cell (PBC) images dataset containing samples of eight different cell types: neutrophils, eosinophils, basophils, lymphocytes, monocytes, IG, erythroblasts, and platelets \cite{ACEVEDO2020105474}. Our work provides state-of-the-art results on the PBC classification task without the need of manual feature extraction or designing complex and hybrid architectures. The main contributions of this paper are as follows:
\begin{enumerate}
    \item Train and evaluate an end-to-end deep learning-based classification system to recognize eight different blood cell types in peripheral blood smear.
    \item Explore and benchmark 27 standard deep CNN architectures for blood cell classification using transfer learning.
    \item Exploit data augmentation and ensembling techniques to further improve the model performance. Our model achieves state-of-the-art performance over previously published works.
\end{enumerate}

\section{Related Work}
\label{related-work}

Computer-aided PBC classification is a challenging problem, and it has been studied extensively for the last few decades. Its solution can be used to develop tools and software that can assist doctors and radiologists in examining and diagnosing many blood-related diseases. Early attempts at this include designing the handcrafted features by experts, which can be further used to perform classification with the pattern recognition algorithms. The application of morphological analysis, including but not limited to morphological operators for PBC segmentation and classification, is the research area that has been explored in great detail in \cite{kim2001automatic, DIRUBERTO2002133, 1397242, scotti2005automatic, dorini2007white, 4167903, taherisadr2013new, lee2014cell, https://doi.org/10.1111/ijlh.12818} before the deep learning-based approach became popular. There has been a significant increase in the studies on the deep learning-based WBC classification approaches \cite{Su2014ANA, Othman2017, Jiang2018, 8376476, 8990301, 10.1007/978-981-13-3600-3_13,8669049, shahin2019white, Almezhghwi2020ImprovedCO, Baydilli2020, Sahlol2020EfficientCO} in the last decade, especially after the successful application of CNNs on other computer vision problems.  The works \cite{SARASWAT201420, Rawat20151948, al2018classification, Patodia2020ASO}, provide a review of the state-of-the-art methods of leukocyte segmentation, feature extraction, and classification published in the last two decades. The authors identified scope for improvement in several aspects compared to the classical approach involving image preprocessing and manual feature extraction, which achieved low performance and required expertise for labeling and handcrafting features. The usage of deep learning based techniques in the medical field has become popular due to the development of more efficient algorithms \cite{kumar2016ensemble, ravi2016deep, cabitza2018machine}. There has also been a wide array of research in the area of RBC segmentation and classification \cite{TOMARI2014206, 10.1145/3177404.3177438, 10.1371/journal.pcbi.1005746, electronics9030427} for diagnosis of many blood-related diseases. There are many different PBC sub-types; however, most studies in this direction focus only on the major types like leukocyte or RBC. Recently, there are attempts at blood cell analysis, encapsulating all different blood cells classes with the same model \cite{ACEVEDO2019105020, Ucar2020}, and it can be more useful, having a large number of applications. Our models outperform the previously published works by successfully applying various standard deep CNN architectures on the large dataset \cite{ACEVEDO2020105474} of microscopic images of peripheral blood cells.

\section{Methodology}
\label{methodology}
\subsection{Overview}
In this work, our primary goal is to develop a deep learning-based end-to-end system to perform PBC recognition. We apply transfer learning to fine-tune standard state-of-the-art deep CNN architectures pre-trained on the ImageNet dataset for the PBC classification task. This approach leverages the already learned features and fine-tunes the models to learn specialized features of microscopic blood cell images. After fine-tuning these models, we apply ensemble techniques to improve the system's overall performance utilizing the knowledge of all trained models, wherein our voting-based ensemble outperforms all other models and previously published works.

\subsection{CNN Architectures}
Deep CNNs have been demonstrated to outperform all other traditional machine learning algorithms for a variety of computer vision tasks in the last decade. Some of the exciting CNN application areas include Image Classification and Segmentation, Face Recognition, Object Detection, Video Processing, Natural Language Processing, and Speech Recognition. The powerful learning ability of deep CNNs is primarily due to multiple feature extraction stages that can automatically learn representations from the data. The CNN architecture generally consists of a feature extractor - containing several convolutional and pooling layers followed by a classifier - containing one or more fully connected layers where the last layer has softmax activation for classification. Depending upon the type of architectural modifications, CNNs can be broadly divided into seven different categories: spatial exploitation, depth, multi-path, width, feature-map exploitation, channel boosting, and attention-based CNNs \cite{Khan_2020}. We utilize 27 standard CNN architectures and analyze their performance for the blood cell classification task. The details about these models can be found in Appendix \ref{appendix-A}. For our experiments, we use the Python implementation of these architectures available in the PyTorch model zoo.

\subsection{Transfer learning}
In machine learning, transfer learning (TL) is a technique where the knowledge gained solving one problem is applied to a different but related problem \cite{ventura2007theoretical}. In the last few years, several transfer learning techniques have been developed and successfully applied using deep neural networks in domains like computer vision, reinforcement learning and natural language processing \cite{6847217, weiss2016survey, csurka2017comprehensive,tan2018survey,  9134370, taylor2009transfer}. We employ transfer learning to solve the blood cell classification problem by fine-tuning 27 different standard deep CNN architectures pre-trained on the ImageNet dataset. As the size of our dataset is not as huge as ImageNet, transfer learning is a suitable choice for our use-case. We use the feature extractor from the pre-trained models and pass the extracted features through a single layer fully connected classifier with eight output units. The fine-tuning of the feature extractor allows us to take advantage of the learned knowledge to further learn new specialized features from blood cell images. This makes the end-to-end training process easier and faster compared to training the whole model from scratch.

\section{Experiments}
This section presents the details of the experiments performed along with the results for peripheral blood cell classification. All the experiments were performed using Google's Colaboratory platform with GPU backend. The code is written in Python using Pytorch deep learning framework.
\label{experiments}
\subsection{Dataset}
\begin{figure}
\begin{floatrow}
\ffigbox{%
\begin{center}
\centerline{\includegraphics[width=0.85\columnwidth]{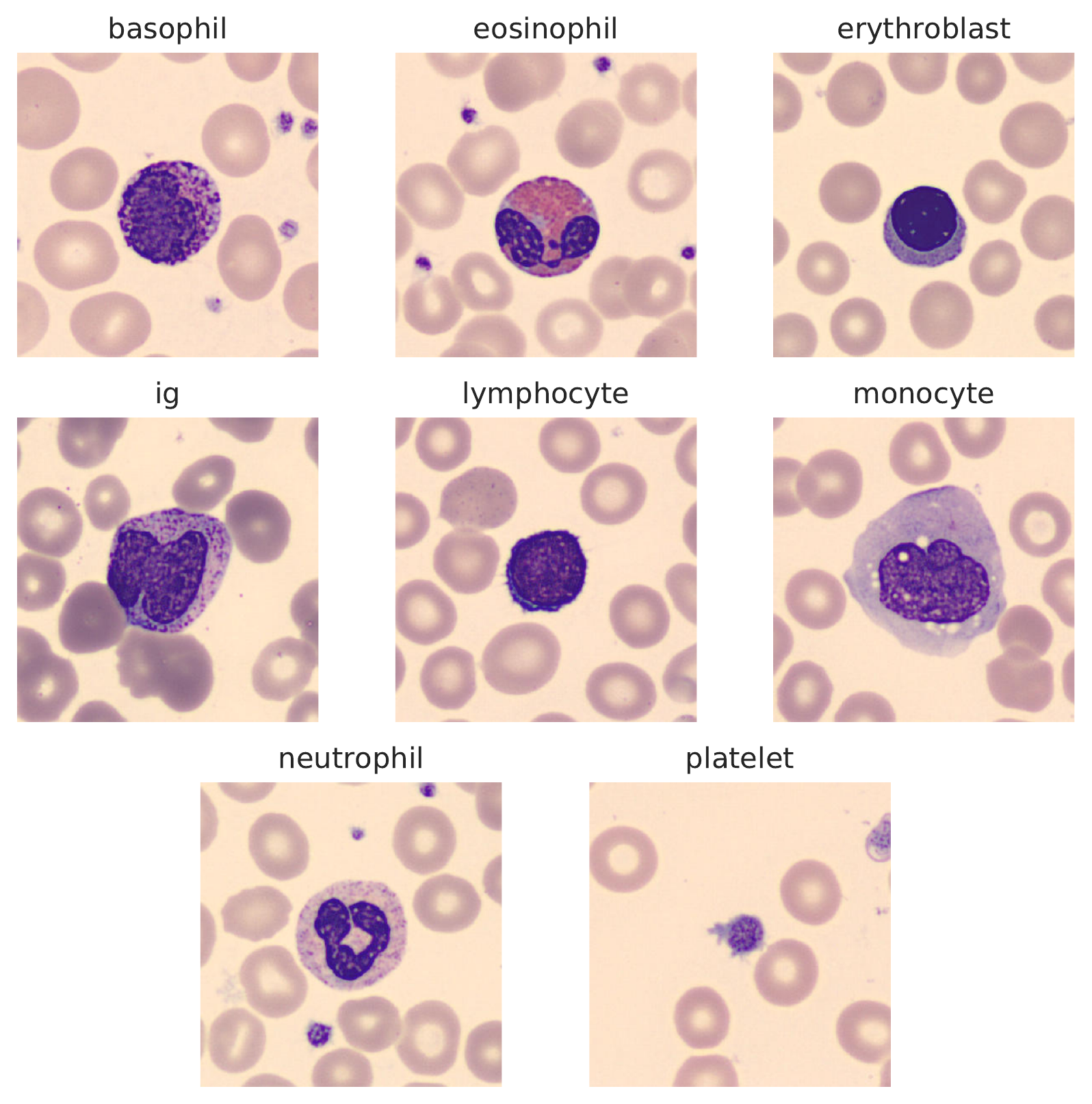}}
\end{center}
\vskip -0.45in
}{%
  \caption{Sample images from PBC dataset}%
  \label{dataset-img}
}
\capbtabbox{%
\begin{small}
\begin{tabular}{lccr}
\toprule
Cell Type & \#Images & Percent(\%) \\
\midrule
Neutrophils & 3329 & 19.48 \\
Eosinophils & 3117 & 18.24 \\
Basophils & 1218 & 7.13 \\
Lymphocytes & 1214 & 7.1 \\
Monocytes & 1420 & 8.31 \\
Immature  & 2895 & 16.94 \\
granulocytes(ig) & & \\
Erythroblasts & 1551 & 9.07 \\
Platelets & 2348 & 13.74 \\
\midrule
Total & 17092 & 100 \\
\bottomrule
\end{tabular}
\end{small}
}{%
  \caption{Cell types and number of samples in each class}%
  \label{tab:dataset-table}
}
\end{floatrow}
\vskip -0.2in
\end{figure}

For the classification of blood cells, peripheral blood cells dataset \cite{ACEVEDO2020105474} was used which contains a total of 17,092 RGB images of normal blood cells, acquired using the analyser CellaVision DM96. The images are 360x363 dimensions each and were labelled by expert clinical pathologists at the Hospital Clinic of Barcelona. The traceability to the patient data was removed, resulting in an anonymized dataset. It is to be noted that the subjects were individuals without infection, hematologic or oncologic disease and free of any pharmacologic treatment at the moment of blood collection. No further filtering or image processing techniques were performed to the images in the dataset. Figure \ref{dataset-img} shows a few sample images from the dataset. Number of instances in each class are presented in the Table \ref{tab:dataset-table}. We split the dataset retaining 64\% samples in each class as training samples, 24\% as validation, and the rest 12\% as testing samples while maintaining the sample per class ratio in all the sets as the original dataset.

\subsection{Training Configuration}

We perform our experiments with 27 different deep CNN models pre-trained on the ImageNet dataset. We use the stochastic gradient descent optimizer with momentum of 0.9, and adopt the weight initialization for classifier as in \cite{7410480}. These models are trained with a minibatch size of 32 on 1 GPU. We start with a learning rate of 0.001 and decay it by a factor of 0.1 every 7 epochs, and terminate training at 25 iterations as model converges. We use data augmentation techniques during training to deal with the limited amount of data which would help the model to generalize well on the unseen samples. The training set is augmented using common image transformations like random vertical and horizontal flips and rotating image by 60 degrees. At this stage, the images are also resized and centered-cropped to dimensions 299x299 for InceptionV3 model and 224x224 for rest of the architectures. For testing, we only evaluate the single view of the original 360×363 image. We use the softmax activation function at the output layer of the classifier and categorical cross-entropy as a loss function. We evaluate the performance of our models by calculating the confusion matrix and reporting evaluation metrics like accuracy, precision, sensitivity and specificity. The formulae of the parameters can be found in Appendix \ref{appendix-B}.

\subsection{Results}
Table \ref{tab:accuracy-table} presents the class-wise and overall test accuracy metrics for all the models.  The four top performing models are highlighted in bold which we use further for ensembling. We observe a minute difference (\textless 1\%) in overall accuracies of all the models. The lowest accuracy of 98.39\% is reported on AlexNet and the highest accuracy is reported as 99.32\% on Wide ResNet-50-2. The confusion matrices and training plots for the top models are presented in Appendix \ref{appendix-C}.

\begin{figure}[!ht]
\begin{center}
\centerline{\includegraphics[width=0.65\columnwidth]{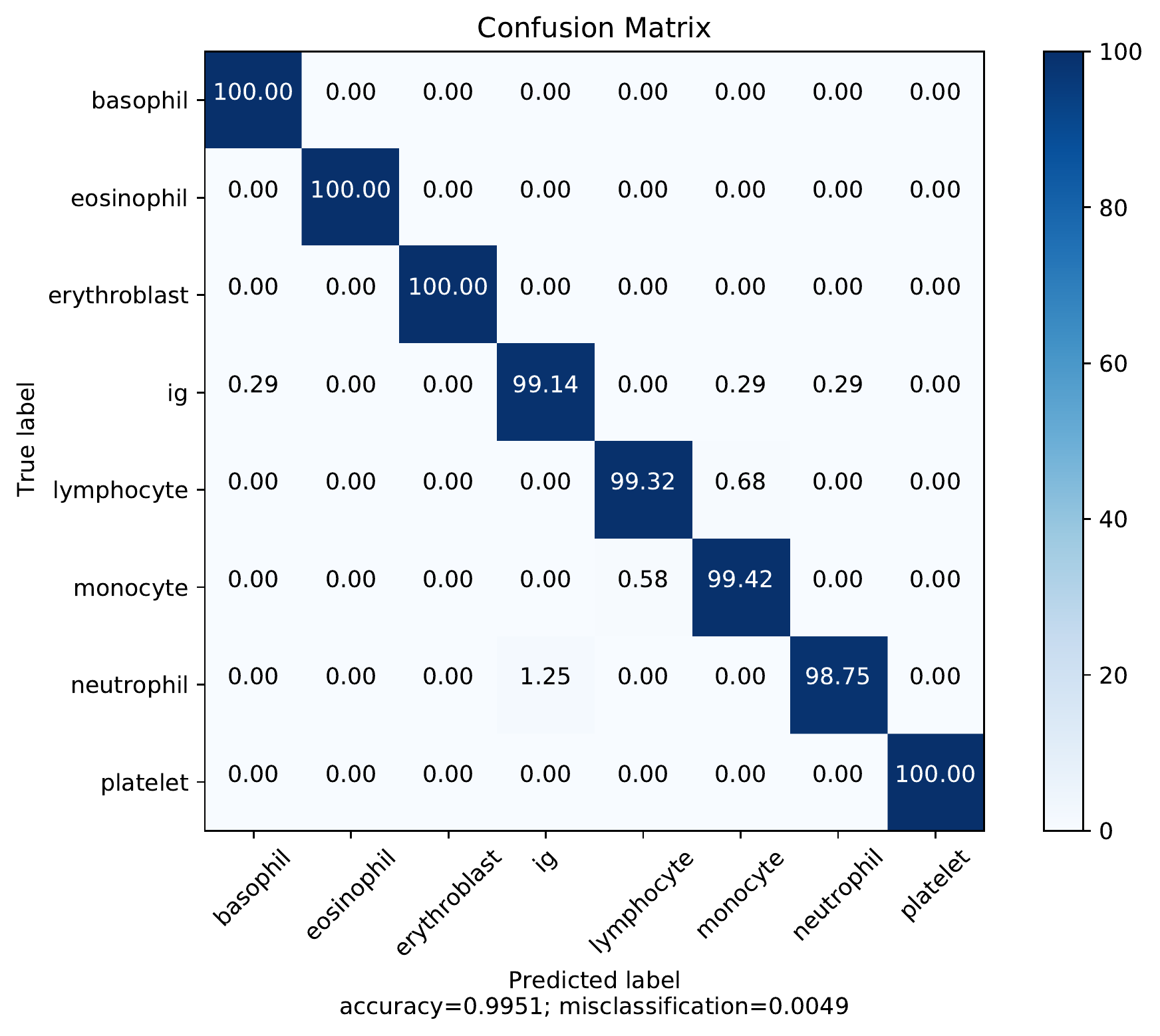}}
\caption{Confusion matrix for the Ensemble model on the test data}
\label{confusion-mat}
\end{center}
\vskip -0.5in
\end{figure}

To further improve the performance, we employ voting-based ensemble by combining the top performing four models namely, Wide ResNet-50-2, VGG-19, Wide ResNet-101-2, and ResNet-34. For each instance in the test set, the class predicted by majority of the four classifiers is considered. This method improves the overall classification performance, achieving 99.51\% overall accuracy, compared to the best model, Wide ResNet-50-2. It should be noted that there are five other models with overall accuracy equal to that of ResNet-34 that is, 99.17\%, but we pick ResNet-34 over them because the validation performance of ensemble is highest with ResNet-34. Figure \ref{confusion-mat} shows the confusion matrix for the ensemble model. Table \ref{prev-work-compare} reports the accuracy, precision, sensitivity and specificity metrics denoting that our ensemble outperforms the previously published works as well as the top performing models.

\begin{table*}[t]
\label{accuracy-table}
\centering
\setlength
\tabcolsep{2pt}
\floatconts
  {tab:accuracy-table}%
  {\caption{Class-wise \& Overall Test Accuracy values after fine-tuning various Deep CNN models for Peripheral Blood Cell Classification Task}}%
{
\renewcommand{\arraystretch}{1.0}
\resizebox{\columnwidth}{!}{
\begin{tabular}{|c|cccccccc|c|}
\hline
\bfseries Architecture & \bfseries Basophil & \bfseries Eosinophil & \bfseries Erythroblast & \bfseries IG & \bfseries Lymphocyte & \bfseries Monocyte & \bfseries Neutrophil & \bfseries Platelet & \bfseries Overall \\
\hline
AlexNet & 0.9864 & 0.9973 & 0.9893 & 0.9483 & 0.9932 & 0.9766 & 0.9875 & 1.0000 & 0.9839 \\
\hline
Densenet-121 & 0.9932 & 0.9973 & 0.9893 & 0.9713 & 0.9932 & 0.9942 & 0.9850 & 1.0000 & 0.9893 \\
Densenet-161 & 1.0000 & 0.9973 & 0.9947 & 0.9770 & 0.9863 & 0.9883 & 0.9775 & 1.0000 & 0.9888 \\
Densenet-169 & 1.0000 & 0.9973 & 0.9893 & 0.9885 & 0.9863 & 0.9942 & 0.9775 & 1.0000 & 0.9908 \\
Densenet-201 & 0.9864 & 0.9973 & 0.9947 & 0.9885 & 0.9863 & 0.9942 & 0.9775 & 1.0000 & 0.9903 \\
\hline
VGG-11bn & 0.9932 & 0.9973 & 0.9786 & 0.9799 & 0.9932 & 0.9825 & 0.9800 & 1.0000 & 0.9878 \\
VGG-11 & 1.0000 & 0.9947 & 0.9947 & 0.9684 & 0.9932 & 0.9708 & 0.9875 & 1.0000 & 0.9878 \\
VGG-13 & 1.0000 & 0.9973 & 0.9893 & 0.9770 & 0.9932 & 0.9708 & 0.9875 & 1.0000 & 0.9893 \\
VGG-13bn & 1.0000 & 0.9973 & 0.9893 & 0.9914 & 0.9932 & 0.9883 & 0.9800 & 1.0000 & 0.9917 \\
VGG-16bn & 1.0000 & 1.0000 & 0.9947 & 0.9856 & 0.9863 & 0.9825 & 0.9725 & 1.0000 & 0.9893 \\
VGG-16 & 0.9932 & 0.9973 & 0.9786 & 0.9828 & 0.9932 & 0.9825 & 0.9775 & 0.9965 & 0.9874 \\
\textbf{VGG-19} & 1.0000 & 1.0000 & 1.0000 & 0.9799 & 0.9932 & 0.9883 & 0.9875 & 1.0000 & \textbf{0.9927} \\
VGG-19bn & 0.9864 & 0.9973 & 0.9893 & 0.9828 & 0.9795 & 0.9883 & 0.9775 & 1.0000 & 0.9878 \\
\hline
ResNet-18 & 1.0000 & 1.0000 & 0.9947 & 0.9856 & 0.9932 & 0.9883 & 0.9800 & 1.0000 & 0.9917 \\
\textbf{ResNet-34} & 1.0000 & 0.9973 & 0.9947 & 0.9885 & 0.9932 & 0.9883 & 0.9800 & 1.0000 & \textbf{0.9917} \\
ResNet-50 & 1.0000 & 0.9973 & 0.9947 & 0.9856 & 0.9932 & 0.9942 & 0.9775 & 1.0000 & 0.9912 \\
ResNet-101 & 0.9932 & 1.0000 & 1.0000 & 0.9828 & 0.9863 & 0.9825 & 0.9875 & 1.0000 & 0.9917 \\
ResNet-152 & 1.0000 & 0.9973 & 0.9893 & 0.9856 & 0.9863 & 0.9942 & 0.9750 & 1.0000 & 0.9898 \\
\hline
ResNeXt-50-32x4d & 0.9932 & 1.0000 & 0.9947 & 0.9770 & 0.9863 & 0.9942 & 0.9825 & 1.0000 & 0.9903 \\
ResNeXt-101-32x8d & 0.9864 & 0.9973 & 1.0000 & 0.9914 & 0.9863 & 0.9942 & 0.9800 & 1.0000 & 0.9917 \\
\hline
SqueezeNet 1.0 & 0.9932 & 0.9973 & 0.9893 & 0.9799 & 0.9932 & 0.9708 & 0.9825 & 1.0000 & 0.9883 \\
SqueezeNet 1.1 & 0.9932 & 0.9973 & 0.9786 & 0.9540 & 0.9932 & 0.9591 & 0.9925 & 1.0000 & 0.9839 \\
\hline
\textbf{Wide ResNet-50-2} & 0.9932 & 1.0000 & 0.9947 & 0.9856 & 0.9932 & 0.9942 & 0.9875 & 1.0000 & \textbf{0.9932} \\
\textbf{Wide ResNet-101-2} & 1.0000 & 1.0000 & 1.0000 & 0.9799 & 0.9932 & 0.9942 & 0.9825 & 1.0000 & \textbf{0.9922} \\
\hline
GoogLeNet & 1.0000 & 0.9973 & 0.9947 & 0.9856 & 0.9932 & 0.9825 & 0.9775 & 0.9965 & 0.9898 \\
\hline
Inception-v3 & 1.0000 & 0.9973 & 1.0000 & 0.9828 & 0.9932 & 0.9825 & 0.9850 & 1.0000 & 0.9917 \\
\hline
MobileNet-v2 & 0.9932 & 0.9973 & 1.0000 & 0.9856 & 0.9932 & 0.9883 & 0.9800 & 1.0000 & 0.9912 \\
\hline
\end{tabular}}}
\vskip -0.1in
\end{table*}

\begin{table}[!ht]
\caption{Comparison of Overall Classification Performance metrics (in \%) of previous published works with our approach on test set of PBC image dataset}
\label{prev-work-compare}
\begin{center}
\begin{small}
\begin{tabular}{lccccr}
\toprule
Method &  Accuracy &  Precision &  Sensitivity &  Specificity\\
\midrule
\cite{ACEVEDO2019105020} & 96.20 & 97.00 & 96.00 & 97.00 \\
\cite{Ucar2020} & 97.94 & 97.94 & 97.94 & 99.71 \\
\cite{long2021bloodcaps} & 99.30 & 99.17 & 99.16 & 99.88\\
\midrule
ResNet-34 (Ours) & 99.17 & 99.20 & 99.27 & 99.88 \\
Wide ResNet-101-2 (Ours) & 99.22 & 99.24 & 99.37 & 99.89 \\
VGG-19 (Ours) & 99.27 & 99.17 & 99.36 & 99.89 \\
Wide ResNet-50-2 (Ours) & 99.32 & 99.29 & 99.35 & 99.90 \\
\midrule
\bfseries Ensemble model (Ours) & \bfseries 99.51 & \bfseries 99.47 & \bfseries 99.58 & \bfseries 99.93 \\
\bottomrule
\end{tabular}
\end{small}
\end{center}
\vskip -0.3in
\end{table}

\section{Discussion}
\label{discussion}
The automatic peripheral blood cell classification is an interesting problem studied widely in recent decades, which can help in the diagnosis of several blood-related diseases. Early attempts to solve this problem involved applying traditional machine learning algorithms with manual feature extraction and morphological analysis. In recent years, deep learning techniques, especially CNNs are employed extensively in this direction. In this work, we solve this problem using deep CNNs, wherein we finetune 27 standard CNN architectures pre-trained on the ImageNet dataset. We further improve the overall performance using the voting-based ensemble of the top four models. Table \ref{prev-work-compare} shows the comparison of our work with the previously published works where our ensemble model outperforms all of them, achieving a classification accuracy of 99.51\%. 
It is worth noting that our model achieves a true positive rate (TPR) of 100\% on basophils, eosinophil, erythroblasts, and platelets, 99.42\% on monocytes, 99.32\% on lymphocytes, 99.14\% on IG, and 98.75\% on neutrophil. From the confusion matrix in Figure \ref{confusion-mat}, we can see that 1.25\% of neutrophils are classified as IG. This may be due to the morphological similarities between the two classes, mainly in the nucleus shape. Moreover, single instances, each from classes lymphocyte and monocyte, are misclassified, which may be the anomalies in dataset.

In addition to our ensemble model, our finetuned CNN models AlexNet (98.39\%), ResNet-18 (99.17\%), VGG-16 (98.74\%), InceptionV3 (99.17\%) show considerable improvement in performance over the corresponding models in the past related works. \cite{ACEVEDO2019105020} achieve overall accuracy 96.2\% on VGG-16 and 95\% on InceptionV3 and \cite{long2021bloodcaps} achieve overall accuracies of 81.5\%, 95.9\%, 97.8\% and 98.4\% on AlexNet, ResNet-18, VGG-16 and InceptionV3 respectively, which is considerably lower than our reported values. Such a significant difference in performance may be due to our careful selection of hyperparameters and weight initialization. Moreover, we notice that even with different architectures with varying configurations of depth, number of parameters, layers, or branches, models don't show a vast difference in the overall performance. The reason behind it may be the simplicity and the smaller size of the PBC dataset compared to the ImageNet which has 1000 classes with over 14 million images making it a much harder dataset wherein the changes in architecture produce notable performance differences. It is noteworthy that the standard CNN architectures we fine-tuned, outperform the specialized architectures and techniques designed \cite{Almezhghwi2020ImprovedCO,long2021bloodcaps} for the peripheral blood cell classification.
\section{Conclusion}
\label{conclusion}
In this work, we solve peripheral blood cell classification problem by fine-tuning standard deep CNN architectures pre-trained on ImageNet dataset. We employ transfer learning to utilize features already learned by state-of-the-art deep learning models and fine-tune them to learn new specialized features of blood cells. In our experiments, all 27 models achieve \(\ge\) 98\% accuracy and 14 of them achieve \(\ge\) 99\% accuracy, showing significant improvements over past works. The Wide ResNet-50-2 achieves the highest accuracy of 99.32\% and the VGG-19 achieves the second highest accuracy of 99.27\%. An ensemble of the top performing models outperforms all other models, achieving state-of-the-art results with a classification accuracy of 99.51\%. Our work provides an empirical baselines for deep learning-based blood cell classification against which the future works with specialized architectures and techniques can be compared. We think that our findings will help further advances in research in deep learning-based automatic blood cell classification.

\bibliography{references}

\begin{thebibliography}{72}
\providecommand{\natexlab}[1]{#1}
\providecommand{\url}[1]{\texttt{#1}}
\expandafter\ifx\csname urlstyle\endcsname\relax
  \providecommand{\doi}[1]{doi: #1}\else
  \providecommand{\doi}{doi: \begingroup \urlstyle{rm}\Url}\fi

\bibitem[Acevedo et~al.(2019)Acevedo, Alférez, Merino, Puigví, and
  Rodellar]{ACEVEDO2019105020}
Andrea Acevedo, Santiago Alférez, Anna Merino, Laura Puigví, and José
  Rodellar.
\newblock Recognition of peripheral blood cell images using convolutional
  neural networks.
\newblock \emph{Computer Methods and Programs in Biomedicine}, 180:\penalty0
  105020, 2019.
\newblock ISSN 0169-2607.
\newblock \doi{https://doi.org/10.1016/j.cmpb.2019.105020}.
\newblock URL
  \url{https://www.sciencedirect.com/science/article/pii/S0169260719303578}.

\bibitem[Acevedo et~al.(2020)Acevedo, Merino, Alférez, Ángel Molina, Boldú,
  and Rodellar]{ACEVEDO2020105474}
Andrea Acevedo, Anna Merino, Santiago Alférez, Ángel Molina, Laura Boldú,
  and José Rodellar.
\newblock A dataset of microscopic peripheral blood cell images for development
  of automatic recognition systems.
\newblock \emph{Data in Brief}, 30:\penalty0 105474, 2020.
\newblock ISSN 2352-3409.
\newblock \doi{https://doi.org/10.1016/j.dib.2020.105474}.
\newblock URL
  \url{https://www.sciencedirect.com/science/article/pii/S2352340920303681}.

\bibitem[Al-Dulaimi et~al.(2018)Al-Dulaimi, Banks, Chandran, Tomeo-Reyes, and
  Nguyen~Thanh]{al2018classification}
Khamael Abbas~Khudhair Al-Dulaimi, Jasmine Banks, Vinod Chandran, Inmaculada
  Tomeo-Reyes, and Kien Nguyen~Thanh.
\newblock Classification of white blood cell types from microscope images:
  Techniques and challenges.
\newblock \emph{Microscopy science: Last approaches on educational programs and
  applied research (Microscopy Book Series, 8)}, pages 17--25, 2018.

\bibitem[Almezhghwi and Serte(2020)]{Almezhghwi2020ImprovedCO}
Khaled Almezhghwi and S.~Serte.
\newblock Improved classification of white blood cells with the generative
  adversarial network and deep convolutional neural network.
\newblock \emph{Computational Intelligence and Neuroscience}, 2020, 2020.

\bibitem[Alzubaidi et~al.(2020)Alzubaidi, Fadhel, Al-Shamma, Zhang, and
  Duan]{electronics9030427}
Laith Alzubaidi, Mohammed~A. Fadhel, Omran Al-Shamma, Jinglan Zhang, and
  Ye~Duan.
\newblock Deep learning models for classification of red blood cells in
  microscopy images to aid in sickle cell anemia diagnosis.
\newblock \emph{Electronics}, 9\penalty0 (3), 2020.
\newblock ISSN 2079-9292.
\newblock \doi{10.3390/electronics9030427}.
\newblock URL \url{https://www.mdpi.com/2079-9292/9/3/427}.

\bibitem[{Banik} et~al.(2019){Banik}, {Saha}, and {Kim}]{8669049}
P.~P. {Banik}, R.~{Saha}, and K.~{Kim}.
\newblock Fused convolutional neural network for white blood cell image
  classification.
\newblock In \emph{2019 International Conference on Artificial Intelligence in
  Information and Communication (ICAIIC)}, pages 238--240, 2019.
\newblock \doi{10.1109/ICAIIC.2019.8669049}.

\bibitem[Baydilli and Atila(2020)]{Baydilli2020}
Y.Y. Baydilli and Ü. Atila.
\newblock Classification of white blood cells using capsule networks.
\newblock \emph{Computerized Medical Imaging and Graphics}, 80, 2020.
\newblock \doi{10.1016/j.compmedimag.2020.101699}.

\bibitem[Cabitza and Banfi(2018)]{cabitza2018machine}
Federico Cabitza and Giuseppe Banfi.
\newblock Machine learning in laboratory medicine: waiting for the flood?
\newblock \emph{Clinical Chemistry and Laboratory Medicine (CCLM)}, 56\penalty0
  (4):\penalty0 516--524, 2018.

\bibitem[Chen et~al.(2015)Chen, Wilson, Tyree, Weinberger, and
  Chen]{chen2015compressing}
Wenlin Chen, James~T. Wilson, Stephen Tyree, Kilian~Q. Weinberger, and Yixin
  Chen.
\newblock Compressing neural networks with the hashing trick, 2015.

\bibitem[Constantino and Cogionis(2000)]{constantino2000nucleated}
Benie~T Constantino and Bessie Cogionis.
\newblock Nucleated rbcs—significance in the peripheral blood film.
\newblock \emph{Laboratory Medicine}, 31\penalty0 (4):\penalty0 223--229, 2000.

\bibitem[Csurka(2017)]{csurka2017comprehensive}
Gabriela Csurka.
\newblock A comprehensive survey on domain adaptation for visual applications.
\newblock \emph{Domain adaptation in computer vision applications}, pages
  1--35, 2017.

\bibitem[{Di Ruberto} et~al.(2002){Di Ruberto}, Dempster, Khan, and
  Jarra]{DIRUBERTO2002133}
Cecilia {Di Ruberto}, Andrew Dempster, Shahid Khan, and Bill Jarra.
\newblock Analysis of infected blood cell images using morphological operators.
\newblock \emph{Image and Vision Computing}, 20\penalty0 (2):\penalty0 133 --
  146, 2002.
\newblock ISSN 0262-8856.
\newblock \doi{https://doi.org/10.1016/S0262-8856(01)00092-0}.
\newblock URL
  \url{http://www.sciencedirect.com/science/article/pii/S0262885601000920}.

\bibitem[Dorini et~al.(2007)Dorini, Minetto, and Leite]{dorini2007white}
Leyza~Baldo Dorini, Rodrigo Minetto, and Neucimar~Jer{\^o}nimo Leite.
\newblock White blood cell segmentation using morphological operators and
  scale-space analysis.
\newblock In \emph{XX Brazilian Symposium on Computer Graphics and Image
  Processing (SIBGRAPI 2007)}, pages 294--304. IEEE, 2007.

\bibitem[Fathima and Syeda(2017)]{fathima17}
Fathima and Syeda.
\newblock Blood cells and leukocyte culture - a short review.
\newblock \emph{Open Access Blood Research and Transfusion Journal}, 1, 05
  2017.
\newblock \doi{10.19080/OABTJ.2017.01.555559}.

\bibitem[Frosst and Hinton(2017)]{frosst2017distilling}
Nicholas Frosst and Geoffrey Hinton.
\newblock Distilling a neural network into a soft decision tree.
\newblock \emph{arXiv preprint arXiv:1711.09784}, 2017.

\bibitem[Han et~al.(2015)Han, Mao, and Dally]{han2015deep}
Song Han, Huizi Mao, and William~J Dally.
\newblock Deep compression: Compressing deep neural networks with pruning,
  trained quantization and huffman coding.
\newblock \emph{arXiv preprint arXiv:1510.00149}, 2015.

\bibitem[{He} et~al.(2015){He}, {Zhang}, {Ren}, and {Sun}]{7410480}
K.~{He}, X.~{Zhang}, S.~{Ren}, and J.~{Sun}.
\newblock Delving deep into rectifiers: Surpassing human-level performance on
  imagenet classification.
\newblock In \emph{2015 IEEE International Conference on Computer Vision
  (ICCV)}, pages 1026--1034, 2015.
\newblock \doi{10.1109/ICCV.2015.123}.

\bibitem[He et~al.(2016)He, Zhang, Ren, and Sun]{he2016deep}
Kaiming He, Xiangyu Zhang, Shaoqing Ren, and Jian Sun.
\newblock Deep residual learning for image recognition.
\newblock In \emph{Proceedings of the IEEE conference on computer vision and
  pattern recognition}, pages 770--778, 2016.

\bibitem[Holinstat(2017)]{holinstat2017normal}
Michael Holinstat.
\newblock Normal platelet function.
\newblock \emph{Cancer and Metastasis Reviews}, 36\penalty0 (2):\penalty0
  195--198, 2017.

\bibitem[Huang et~al.(2017)Huang, Liu, Van Der~Maaten, and
  Weinberger]{huang2017densely}
Gao Huang, Zhuang Liu, Laurens Van Der~Maaten, and Kilian~Q Weinberger.
\newblock Densely connected convolutional networks.
\newblock In \emph{Proceedings of the IEEE conference on computer vision and
  pattern recognition}, pages 4700--4708, 2017.

\bibitem[Iandola et~al.(2016)Iandola, Han, Moskewicz, Ashraf, Dally, and
  Keutzer]{iandola2016squeezenet}
Forrest~N Iandola, Song Han, Matthew~W Moskewicz, Khalid Ashraf, William~J
  Dally, and Kurt Keutzer.
\newblock Squeezenet: Alexnet-level accuracy with 50x fewer parameters and< 0.5
  mb model size.
\newblock \emph{arXiv preprint arXiv:1602.07360}, 2016.

\bibitem[Jiang et~al.(2018)Jiang, Cheng, Qin, Du, and Zhang]{Jiang2018}
Ming Jiang, Liu Cheng, Feiwei Qin, Lian Du, and Min Zhang.
\newblock White blood cells classification with deep convolutional neural
  networks.
\newblock \emph{International Journal of Pattern Recognition and Artificial
  Intelligence}, 32, 02 2018.
\newblock \doi{10.1142/S0218001418570069}.

\bibitem[Khan et~al.(2020{\natexlab{a}})Khan, Sohail, and Ali]{khan2020new}
Asifullah Khan, Anabia Sohail, and Amna Ali.
\newblock A new channel boosted convolutional neural network using transfer
  learning, 2020{\natexlab{a}}.

\bibitem[Khan et~al.(2020{\natexlab{b}})Khan, Sohail, Zahoora, and
  Qureshi]{Khan_2020}
Asifullah Khan, Anabia Sohail, Umme Zahoora, and Aqsa~Saeed Qureshi.
\newblock A survey of the recent architectures of deep convolutional neural
  networks.
\newblock \emph{Artificial Intelligence Review}, 53\penalty0 (8):\penalty0
  5455–5516, Apr 2020{\natexlab{b}}.
\newblock ISSN 1573-7462.
\newblock \doi{10.1007/s10462-020-09825-6}.
\newblock URL \url{http://dx.doi.org/10.1007/s10462-020-09825-6}.

\bibitem[Kim et~al.(2001)Kim, Jeon, Choi, Kim, and Ho]{kim2001automatic}
Kyungsu Kim, Jeonghee Jeon, WanKyoo Choi, Pankoo Kim, and Yo-Sung Ho.
\newblock Automatic cell classification in human’s peripheral blood images
  based on morphological image processing.
\newblock In \emph{Australian Joint Conference on Artificial Intelligence},
  pages 225--236. Springer, 2001.

\bibitem[Krizhevsky(2014)]{krizhevsky2014one}
Alex Krizhevsky.
\newblock One weird trick for parallelizing convolutional neural networks.
\newblock \emph{arXiv preprint arXiv:1404.5997}, 2014.

\bibitem[Kuhn et~al.(2017)Kuhn, Diederich, Keller, Kramer, Lückstädt,
  Panknin, Suvorava, Isakson, Kelm, and Cortese-Krott]{kuhn17}
Viktoria Kuhn, Lukas Diederich, T.C.~Stevenson Keller, Christian~M. Kramer,
  Wiebke Lückstädt, Christina Panknin, Tatsiana Suvorava, Brant~E. Isakson,
  Malte Kelm, and Miriam~M. Cortese-Krott.
\newblock Red blood cell function and dysfunction: Redox regulation, nitric
  oxide metabolism, anemia.
\newblock \emph{Antioxidants \& Redox Signaling}, 26\penalty0 (13):\penalty0
  718--742, 2017.
\newblock \doi{10.1089/ars.2016.6954}.
\newblock URL \url{https://doi.org/10.1089/ars.2016.6954}.
\newblock PMID: 27889956.

\bibitem[Kumar et~al.(2016)Kumar, Kim, Lyndon, Fulham, and
  Feng]{kumar2016ensemble}
Ashnil Kumar, Jinman Kim, David Lyndon, Michael Fulham, and Dagan Feng.
\newblock An ensemble of fine-tuned convolutional neural networks for medical
  image classification.
\newblock \emph{IEEE journal of biomedical and health informatics}, 21\penalty0
  (1):\penalty0 31--40, 2016.

\bibitem[L.(2005)]{deanl05}
Dean L.
\newblock Blood groups and red cell antigens, chapter 1, blood and the cells it
  contains., 2005.
\newblock URL \url{https://www.ncbi.nlm.nih.gov/books/NBK2263/}.

\bibitem[Lee and Chen(2014)]{lee2014cell}
Howard Lee and Yi-Ping~Phoebe Chen.
\newblock Cell morphology based classification for red cells in blood smear
  images.
\newblock \emph{Pattern Recognition Letters}, 49:\penalty0 155--161, 2014.

\bibitem[Long et~al.(2021)Long, Peng, Song, Xia, and Sang]{long2021bloodcaps}
Fei Long, Jing-Jie Peng, Weitao Song, Xiaobo Xia, and Jun Sang.
\newblock Bloodcaps: A capsule network based model for the multiclassification
  of human peripheral blood cells.
\newblock \emph{Computer Methods and Programs in Biomedicine}, page 105972,
  2021.

\bibitem[{Macawile} et~al.(2018){Macawile}, {Quiñones}, {Ballado}, {Cruz}, and
  {Caya}]{8376476}
M.~J. {Macawile}, V.~V. {Quiñones}, A.~{Ballado}, J.~D. {Cruz}, and M.~V.
  {Caya}.
\newblock White blood cell classification and counting using convolutional
  neural network.
\newblock In \emph{2018 3rd International Conference on Control and Robotics
  Engineering (ICCRE)}, pages 259--263, 2018.
\newblock \doi{10.1109/ICCRE.2018.8376476}.

\bibitem[Mathur et~al.(2013)Mathur, Tripathi, and Kuse]{mathur2013scalable}
Atin Mathur, Ardhendu~S Tripathi, and Manohar Kuse.
\newblock Scalable system for classification of white blood cells from leishman
  stained blood stain images.
\newblock \emph{Journal of pathology informatics}, 4\penalty0 (Suppl), 2013.

\bibitem[Othman et~al.(2017)Othman, Mohammed, and Baban]{Othman2017}
Mazin~Zeki Othman, Thabit Mohammed, and Alaa Baban.
\newblock Neural network classification of white blood cell using microscopic
  images.
\newblock \emph{(IJACSA) International Journal of Advanced Computer Science and
  Applications}, 8:\penalty0 99--104, 06 2017.
\newblock \doi{10.14569/IJACSA.2017.080513}.

\bibitem[Patodia et~al.(2020)Patodia, Nibhasya, and
  Chandraprabha]{Patodia2020ASO}
Chandni Patodia, G.~Nibhasya, and R.~Chandraprabha.
\newblock A survey on automated methods used for wbc classification.
\newblock In \emph{International Journal of Engineering Technology and
  Management Sciences (IJETMS)}, 2020.

\bibitem[Peng et~al.(2015)Peng, Huang, Qi, Zhao, Zhang, Zhao, Yuan, He, and
  Zhang]{peng2015pku}
Yuxin Peng, Xin Huang, Jinwei Qi, Junjie Zhao, Junchao Zhang, Yunzhen Zhao,
  Yuxin Yuan, Xiangteng He, and Jian Zhang.
\newblock Pku-icst at trecvid 2015: Instance search task.
\newblock In \emph{TRECVID}, 2015.

\bibitem[{Piuri} and {Scotti}(2004)]{1397242}
V.~{Piuri} and F.~{Scotti}.
\newblock Morphological classification of blood leucocytes by microscope
  images.
\newblock In \emph{2004 IEEE International Conference onComputational
  Intelligence for Measurement Systems and Applications, 2004. CIMSA.}, pages
  103--108, 2004.
\newblock \doi{10.1109/CIMSA.2004.1397242}.

\bibitem[Rav{\`\i} et~al.(2016)Rav{\`\i}, Wong, Deligianni, Berthelot,
  Andreu-Perez, Lo, and Yang]{ravi2016deep}
Daniele Rav{\`\i}, Charence Wong, Fani Deligianni, Melissa Berthelot, Javier
  Andreu-Perez, Benny Lo, and Guang-Zhong Yang.
\newblock Deep learning for health informatics.
\newblock \emph{IEEE journal of biomedical and health informatics}, 21\penalty0
  (1):\penalty0 4--21, 2016.

\bibitem[Rawat et~al.(2015)Rawat, Bhadauria, Singh, and Virmani]{Rawat20151948}
Jyoti Rawat, HS~Bhadauria, Annapurna Singh, and Jitendra Virmani.
\newblock Review of leukocyte classification techniques for microscopic blood
  images.
\newblock In \emph{2015 2nd International Conference on Computing for
  Sustainable Global Development (INDIACom)}, pages 1948--1954. IEEE, 2015.

\bibitem[Rodellar et~al.(2018)Rodellar, Alférez, Acevedo, Molina, and
  Merino]{https://doi.org/10.1111/ijlh.12818}
J.~Rodellar, S.~Alférez, A.~Acevedo, A.~Molina, and A.~Merino.
\newblock Image processing and machine learning in the morphological analysis
  of blood cells.
\newblock \emph{International Journal of Laboratory Hematology}, 40\penalty0
  (S1):\penalty0 46--53, 2018.
\newblock \doi{https://doi.org/10.1111/ijlh.12818}.
\newblock URL \url{https://onlinelibrary.wiley.com/doi/abs/10.1111/ijlh.12818}.

\bibitem[Roy et~al.(2017)Roy, Conjeti, Sheet, Katouzian, Navab, and
  Wachinger]{roy2017error}
Abhijit~Guha Roy, Sailesh Conjeti, Debdoot Sheet, Amin Katouzian, Nassir Navab,
  and Christian Wachinger.
\newblock Error corrective boosting for learning fully convolutional networks
  with limited data.
\newblock In \emph{International Conference on Medical Image Computing and
  Computer-Assisted Intervention}, pages 231--239. Springer, 2017.

\bibitem[Sahlol et~al.(2020)Sahlol, Kollmannsberger, and
  Ewees]{Sahlol2020EfficientCO}
Ahmed~T. Sahlol, P.~Kollmannsberger, and A.~A. Ewees.
\newblock Efficient classification of white blood cell leukemia with improved
  swarm optimization of deep features.
\newblock \emph{Scientific Reports}, 10, 2020.

\bibitem[Sandler et~al.(2018)Sandler, Howard, Zhu, Zhmoginov, and
  Chen]{sandler2018mobilenetv2}
Mark Sandler, Andrew Howard, Menglong Zhu, Andrey Zhmoginov, and Liang-Chieh
  Chen.
\newblock Mobilenetv2: Inverted residuals and linear bottlenecks.
\newblock In \emph{Proceedings of the IEEE conference on computer vision and
  pattern recognition}, pages 4510--4520, 2018.

\bibitem[Saraswat and Arya(2014)]{SARASWAT201420}
Mukesh Saraswat and K.V. Arya.
\newblock Automated microscopic image analysis for leukocytes identification: A
  survey.
\newblock \emph{Micron}, 65:\penalty0 20 -- 33, 2014.
\newblock ISSN 0968-4328.
\newblock \doi{https://doi.org/10.1016/j.micron.2014.04.001}.
\newblock URL
  \url{http://www.sciencedirect.com/science/article/pii/S0968432814000663}.

\bibitem[Scotti(2005)]{scotti2005automatic}
Fabio Scotti.
\newblock Automatic morphological analysis for acute leukemia identification in
  peripheral blood microscope images.
\newblock In \emph{CIMSA. 2005 IEEE International Conference on Computational
  Intelligence for Measurement Systems and Applications, 2005.}, pages 96--101.
  IEEE, 2005.

\bibitem[Shafique and Tehsin(2018)]{shafique2018computer}
Sarmad Shafique and Samabia Tehsin.
\newblock Computer-aided diagnosis of acute lymphoblastic leukaemia.
\newblock \emph{Computational and mathematical methods in medicine}, 2018,
  2018.

\bibitem[Shahin et~al.(2019)Shahin, Guo, Amin, and Sharawi]{shahin2019white}
Ahmed~Ismail Shahin, Yanhui Guo, Khalid~M Amin, and Amr~A Sharawi.
\newblock White blood cells identification system based on convolutional deep
  neural learning networks.
\newblock \emph{Computer methods and programs in biomedicine}, 168:\penalty0
  69--80, 2019.

\bibitem[{Shao} et~al.(2015){Shao}, {Zhu}, and {Li}]{6847217}
L.~{Shao}, F.~{Zhu}, and X.~{Li}.
\newblock Transfer learning for visual categorization: A survey.
\newblock \emph{IEEE Transactions on Neural Networks and Learning Systems},
  26\penalty0 (5):\penalty0 1019--1034, 2015.
\newblock \doi{10.1109/TNNLS.2014.2330900}.

\bibitem[Sharma et~al.(2019)Sharma, Bhave, and
  Janghel]{10.1007/978-981-13-3600-3_13}
Mayank Sharma, Aishwarya Bhave, and Rekh~Ram Janghel.
\newblock White blood cell classification using convolutional neural network.
\newblock In Jiacun Wang, G.~Ram~Mohana Reddy, V.~Kamakshi Prasad, and
  V.~Sivakumar Reddy, editors, \emph{Soft Computing and Signal Processing},
  pages 135--143, Singapore, 2019. Springer Singapore.
\newblock ISBN 978-981-13-3600-3.

\bibitem[Simonyan and Zisserman(2014)]{simonyan2014very}
Karen Simonyan and Andrew Zisserman.
\newblock Very deep convolutional networks for large-scale image recognition.
\newblock \emph{arXiv preprint arXiv:1409.1556}, 2014.

\bibitem[Srivastava et~al.(2015)Srivastava, Greff, and
  Schmidhuber]{srivastava2015highway}
Rupesh~Kumar Srivastava, Klaus Greff, and J{\"u}rgen Schmidhuber.
\newblock Highway networks.
\newblock \emph{arXiv preprint arXiv:1505.00387}, 2015.

\bibitem[Su et~al.(2014)Su, Cheng, and Wang]{Su2014ANA}
M.~Su, Chun-Yen Cheng, and P.~Wang.
\newblock A neural-network-based approach to white blood cell classification.
\newblock \emph{The Scientific World Journal}, 2014, 2014.

\bibitem[Szegedy et~al.(2015)Szegedy, Liu, Jia, Sermanet, Reed, Anguelov,
  Erhan, Vanhoucke, and Rabinovich]{szegedy2015going}
Christian Szegedy, Wei Liu, Yangqing Jia, Pierre Sermanet, Scott Reed, Dragomir
  Anguelov, Dumitru Erhan, Vincent Vanhoucke, and Andrew Rabinovich.
\newblock Going deeper with convolutions.
\newblock In \emph{Proceedings of the IEEE conference on computer vision and
  pattern recognition}, pages 1--9, 2015.

\bibitem[Szegedy et~al.(2016)Szegedy, Vanhoucke, Ioffe, Shlens, and
  Wojna]{szegedy2016rethinking}
Christian Szegedy, Vincent Vanhoucke, Sergey Ioffe, Jon Shlens, and Zbigniew
  Wojna.
\newblock Rethinking the inception architecture for computer vision.
\newblock In \emph{Proceedings of the IEEE conference on computer vision and
  pattern recognition}, pages 2818--2826, 2016.

\bibitem[Taherisadr et~al.(2013)Taherisadr, Nasirzonouzi, Baradaran, and
  Mehdizade]{taherisadr2013new}
Mojtaba Taherisadr, Mona Nasirzonouzi, Behzad Baradaran, and Alireza Mehdizade.
\newblock New approch to red blood cell classification using morphological
  image processing.
\newblock \emph{Shiraz E-Medical Journal}, 14\penalty0 (1):\penalty0 44--53,
  2013.

\bibitem[Tan et~al.(2018)Tan, Sun, Kong, Zhang, Yang, and Liu]{tan2018survey}
Chuanqi Tan, Fuchun Sun, Tao Kong, Wenchang Zhang, Chao Yang, and Chunfang Liu.
\newblock A survey on deep transfer learning.
\newblock In \emph{International conference on artificial neural networks},
  pages 270--279. Springer, 2018.

\bibitem[Taylor and Stone(2009)]{taylor2009transfer}
Matthew~E Taylor and Peter Stone.
\newblock Transfer learning for reinforcement learning domains: A survey.
\newblock \emph{Journal of Machine Learning Research}, 10\penalty0 (7), 2009.

\bibitem[{Theera-Umpon} and {Dhompongsa}(2007)]{4167903}
N.~{Theera-Umpon} and S.~{Dhompongsa}.
\newblock Morphological granulometric features of nucleus in automatic bone
  marrow white blood cell classification.
\newblock \emph{IEEE Transactions on Information Technology in Biomedicine},
  11\penalty0 (3):\penalty0 353--359, 2007.
\newblock \doi{10.1109/TITB.2007.892694}.

\bibitem[{Throngnumchai} et~al.(2019){Throngnumchai}, {Lomvisai}, {Tantasirin},
  and {Phasukkit}]{8990301}
K.~{Throngnumchai}, P.~{Lomvisai}, C.~{Tantasirin}, and P.~{Phasukkit}.
\newblock Classification of white blood cell using deep convolutional neural
  network.
\newblock In \emph{2019 12th Biomedical Engineering International Conference
  (BMEiCON)}, pages 1--4, 2019.
\newblock \doi{10.1109/BMEiCON47515.2019.8990301}.

\bibitem[Tomari et~al.(2014)Tomari, Zakaria, Jamil, Nor, and
  Fuad]{TOMARI2014206}
Razali Tomari, Wan Nurshazwani~Wan Zakaria, Muhammad Mahadi~Abdul Jamil,
  Faridah~Mohd Nor, and Nik Farhan~Nik Fuad.
\newblock Computer aided system for red blood cell classification in blood
  smear image.
\newblock \emph{Procedia Computer Science}, 42:\penalty0 206 -- 213, 2014.
\newblock ISSN 1877-0509.
\newblock \doi{https://doi.org/10.1016/j.procs.2014.11.053}.
\newblock URL
  \url{http://www.sciencedirect.com/science/article/pii/S1877050914014914}.
\newblock Medical and Rehabilitation Robotics and Instrumentation (MRRI2013).

\bibitem[Tyas et~al.(2017)Tyas, Ratnaningsih, Harjoko, and
  Hartati]{10.1145/3177404.3177438}
Dyah~Aruming Tyas, Tri Ratnaningsih, Agus Harjoko, and Sri Hartati.
\newblock The classification of abnormal red blood cell on the minor
  thalassemia case using artificial neural network and convolutional neural
  network.
\newblock In \emph{Proceedings of the International Conference on Video and
  Image Processing}, ICVIP 2017, page 228–233, New York, NY, USA, 2017.
  Association for Computing Machinery.
\newblock ISBN 9781450353830.
\newblock \doi{10.1145/3177404.3177438}.
\newblock URL \url{https://doi.org/10.1145/3177404.3177438}.

\bibitem[{Ucar}(2020)]{Ucar2020}
F.~{Ucar}.
\newblock Deep learning approach to cell classificatio in human peripheral
  blood.
\newblock In \emph{2020 5th International Conference on Computer Science and
  Engineering (UBMK)}, pages 383--387, 2020.
\newblock \doi{10.1109/UBMK50275.2020.9219480}.

\bibitem[Ventura and Warnick(2007)]{ventura2007theoretical}
Dan Ventura and Sean Warnick.
\newblock A theoretical foundation for inductive transfer.
\newblock \emph{Brigham Young University, College of Physical and Mathematical
  Sciences}, 19, 2007.

\bibitem[Wang et~al.(2017)Wang, Jiang, Qian, Yang, Li, Zhang, Wang, and
  Tang]{wang2017residual}
Fei Wang, Mengqing Jiang, Chen Qian, Shuo Yang, Cheng Li, Honggang Zhang,
  Xiaogang Wang, and Xiaoou Tang.
\newblock Residual attention network for image classification.
\newblock In \emph{Proceedings of the IEEE conference on computer vision and
  pattern recognition}, pages 3156--3164, 2017.

\bibitem[Weiss et~al.(2016)Weiss, Khoshgoftaar, and Wang]{weiss2016survey}
Karl Weiss, Taghi~M Khoshgoftaar, and DingDing Wang.
\newblock A survey of transfer learning.
\newblock \emph{Journal of Big data}, 3\penalty0 (1):\penalty0 1--40, 2016.

\bibitem[Woo et~al.(2018)Woo, Park, Lee, and Kweon]{woo2018cbam}
Sanghyun Woo, Jongchan Park, Joon-Young Lee, and In~So Kweon.
\newblock Cbam: Convolutional block attention module.
\newblock In \emph{Proceedings of the European conference on computer vision
  (ECCV)}, pages 3--19, 2018.

\bibitem[Wu et~al.(2016)Wu, Leng, Wang, Hu, and Cheng]{wu2016quantized}
Jiaxiang Wu, Cong Leng, Yuhang Wang, Qinghao Hu, and Jian Cheng.
\newblock Quantized convolutional neural networks for mobile devices.
\newblock In \emph{Proceedings of the IEEE Conference on Computer Vision and
  Pattern Recognition}, pages 4820--4828, 2016.

\bibitem[Xie et~al.(2017)Xie, Girshick, Doll{\'a}r, Tu, and
  He]{xie2017aggregated}
Saining Xie, Ross Girshick, Piotr Doll{\'a}r, Zhuowen Tu, and Kaiming He.
\newblock Aggregated residual transformations for deep neural networks.
\newblock In \emph{Proceedings of the IEEE conference on computer vision and
  pattern recognition}, pages 1492--1500, 2017.

\bibitem[Xu et~al.(2017)Xu, Papageorgiou, Abidi, Dao, Zhao, and
  Karniadakis]{10.1371/journal.pcbi.1005746}
Mengjia Xu, Dimitrios~P. Papageorgiou, Sabia~Z. Abidi, Ming Dao, Hong Zhao, and
  George~Em Karniadakis.
\newblock A deep convolutional neural network for classification of red blood
  cells in sickle cell anemia.
\newblock \emph{PLOS Computational Biology}, 13\penalty0 (10):\penalty0 1--27,
  10 2017.
\newblock \doi{10.1371/journal.pcbi.1005746}.
\newblock URL \url{https://doi.org/10.1371/journal.pcbi.1005746}.

\bibitem[Zagoruyko and Komodakis(2016)]{zagoruyko2016wide}
Sergey Zagoruyko and Nikos Komodakis.
\newblock Wide residual networks.
\newblock \emph{arXiv preprint arXiv:1605.07146}, 2016.

\bibitem[Zhang et~al.(2018)Zhang, Zhou, Lin, and Sun]{zhang2018shufflenet}
Xiangyu Zhang, Xinyu Zhou, Mengxiao Lin, and Jian Sun.
\newblock Shufflenet: An extremely efficient convolutional neural network for
  mobile devices.
\newblock In \emph{Proceedings of the IEEE conference on computer vision and
  pattern recognition}, pages 6848--6856, 2018.

\bibitem[{Zhuang} et~al.(2021){Zhuang}, {Qi}, {Duan}, {Xi}, {Zhu}, {Zhu},
  {Xiong}, and {He}]{9134370}
F.~{Zhuang}, Z.~{Qi}, K.~{Duan}, D.~{Xi}, Y.~{Zhu}, H.~{Zhu}, H.~{Xiong}, and
  Q.~{He}.
\newblock A comprehensive survey on transfer learning.
\newblock \emph{Proceedings of the IEEE}, 109\penalty0 (1):\penalty0 43--76,
  2021.
\newblock \doi{10.1109/JPROC.2020.3004555}.

\end{thebibliography}

\appendix

\section{Evaluation Metrics Formulae}
\label{appendix-B}
Below are the performance metrics used to evaluate the classification model on PCB dataset.
\begin{align*}
    Overall\: Accuracy = \frac{Total\:Number\:of\:Correct\:Predictions}{Total\:Number\:of\:Samples}
\end{align*}
\begin{align*}
   Precision = \frac{TP}{TP+FP},\: Sensitivity = \frac{TP}{TP+FN},\: Specificity = \frac{TN}{TN+FP}
\end{align*}
where TP (True Positive) is the value of principal diagonal of target class in confusion matrix, TN (True Negative) is the sum of all the elements except the row and column of target class, FP (False Positive) is the sum of all column elements of target class excluding its TP , and FN (False Negative) is the sum of all row elements of target class excluding its TP. For this multi-class classification problem, we considered one-vs-all approach for calculating specificity, sensitivity and precision from the confusion matrix.

\section{More details on CNN Architectures}
\label{appendix-A}
The evolution of the CNN architectures in the last decade has been remarkable. Initially, different ideas such as information gating mechanism across multiple layers, skip connections, and cross-layer channel connectivity were introduced \cite{srivastava2015highway, he2016deep, huang2017densely}. Then, the focus of research shifted mainly on designing of generic blocks that can be inserted at any learning stage in CNN architecture to improve the network representation \cite{peng2015pku} and can be used to assign attention to spatial and feature map information \cite{wang2017residual, roy2017error, woo2018cbam}. In 2018, an idea of channel boosting was introduced by Khan et al. \cite{khan2020new} to boost the performance of a CNN by learning distinct features and utilizing the already learned features using the TL. Some of the salient improvements involve knowledge distillation and compression or squeezing of pre-trained networks \cite{chen2015compressing, han2015deep, wu2016quantized, frosst2017distilling}.
GoogleNet replaced the conventional convolution operation with point-wise group convolution operation to make it small and computationally efficient \cite{szegedy2015going}. ShuffleNet also used the point-wise group convolution but with a new idea of channel shuffle that reduced the number of operations significantly without impacting the performance \cite{zhang2018shufflenet}. For our experiments we used 27 standard CNN architectures, details of which are presented in Table \ref{tab:model_archs}. 

\begin{table*}[!h]
    \centering
    \renewcommand{\arraystretch}{1.2}
    \resizebox{\columnwidth}{!}{
    \begin{tabular}{|c|c|c|c|c|c|}
\hline
\textbf{Architecture} & \textbf{Top-1} & \textbf{Top-5} & \textbf{Depth} & \textbf{\#Params} & \textbf{Main Contribution} \\
& \textbf{Error} & \textbf{Error} & & \textbf{(millions)} & \\
\hline
AlexNet & 43.45 & 20.91 & 5 & 61 & - Deeper and wider than the LeNet \\
\cite{krizhevsky2014one} & & & & & - Uses Relu, dropout and overlap Pooling \\
& & & & & - GPUs NVIDIA GTX 580 \\
\hline
VGG-11 & 30.98 & 11.37 & 11 & 133 & \\
VGG-13 & 30.07 & 10.75 & 13 & 133 & \\
VGG-16 & 28.41 & 9.62 & 16 & 138 & \\
VGG-19 & 27.62 & 9.12 & 19 & 144 & - Analysis with different depths \\
VGG-11 bn & 29.62 & 10.19 & 11 & 133 & - Homogenous topology\\
VGG-13 bn & 28.45 & 9.63 & 13 & 133 & - Uses small size kernels \\
VGG-16 bn & 26.63 & 8.5 & 16 & 138 & \\
VGG-19 bn & 25.76 & 8.15 & 19 & 144 & \\
\cite{simonyan2014very} & & & & & \\
\hline
ResNet-18 & 30.24 & 10.92 & 18 & 11 & \\
ResNet-34 & 26.7 & 8.58 & 34 & 21 & \\
ResNet-50 & 23.85 & 7.13 & 50 & 25 & - Residual learning framework\\
ResNet-101 & 22.63 & 6.44 & 101 & 44 & - Identity mapping based skip connections\\
ResNet-152 & 21.69 & 5.94 & 152 & 60 & - Ease the training of deeper networks\\
\cite{he2016deep} & & & & & \\
\hline
SqueezeNet 1.0 & 41.9 & 19.58 & 18 & 1.24 & - 50x reduction in model size of AlexNet \\
SqueezeNet 1.1 & 41.81 & 19.38 & 18 & 1.23 & - Deep compression with 8-bit quantization \\
\cite{iandola2016squeezenet} & & & & & \\
\hline
Densenet-121 & 25.35 & 7.83 & 121 & 8 & \multirow{1}{*}{ - Cross-layer information flow } \\
Densenet-169 & 24 & 7 & 169 & 14 & - Alleviate the vanishing-gradient problem\\
Densenet-201 & 22.8 & 6.43 & 201 & 20 & - Reduced number of parameters\\
Densenet-161 & 22.35 & 6.2 & 161 & 28 & - Encourage Feature Reuse\\
\cite{huang2017densely} & & & & & \\
\hline
Inception v3 & 22.55 & 6.44 & 48 & 24 & - Handles representational bottleneck \\
\cite{szegedy2016rethinking} & & & & & - Replace large size filters with small filters \\
\hline
GoogleNet & 30.22 & 10.47 & 22 & 6 & - Introduced block concept \\
\cite{szegedy2015going} & & & & & - Split transform and merge idea \\
\hline
MobileNet V2 & 28.12 & 9.71 & 53 & 3 & - Inverted residual structure \\
\cite{sandler2018mobilenetv2} & & & & & - Non-linearities in narrow layers removed \\
\hline
ResNeXt-50-32x4d & 22.38 & 6.3 & 50 & 25 & - Cardinality \\
ResNeXt-101-32x8d & 20.69 & 5.47 & 101 & 88 & - Homogeneous topology \\
\cite{xie2017aggregated} & & & & & - Grouped convolution \\
\hline
Wide ResNet-50-2 & 21.49 & 5.91 & 50 & 68 & - Feature Reuse and Fast training\\
Wide ResNet-101-2 & 21.16 & 5.72 & 101 & 126 & - Width is increased and depth is decreased\\
\cite{zagoruyko2016wide} & & & & & \\
\hline
\end{tabular}}
    \caption{Overview of CNN model architectures. Error rates (\%) on ImageNet dataset as reported in Pytorch documentation.}
    \label{tab:model_archs}
\end{table*}

\section{Additional Evaluation Results}
\label{appendix-C}
\begin{figure}[h]
\begin{floatrow}
\ffigbox{%
\begin{center}
\centerline{\includegraphics[width=0.95\columnwidth]{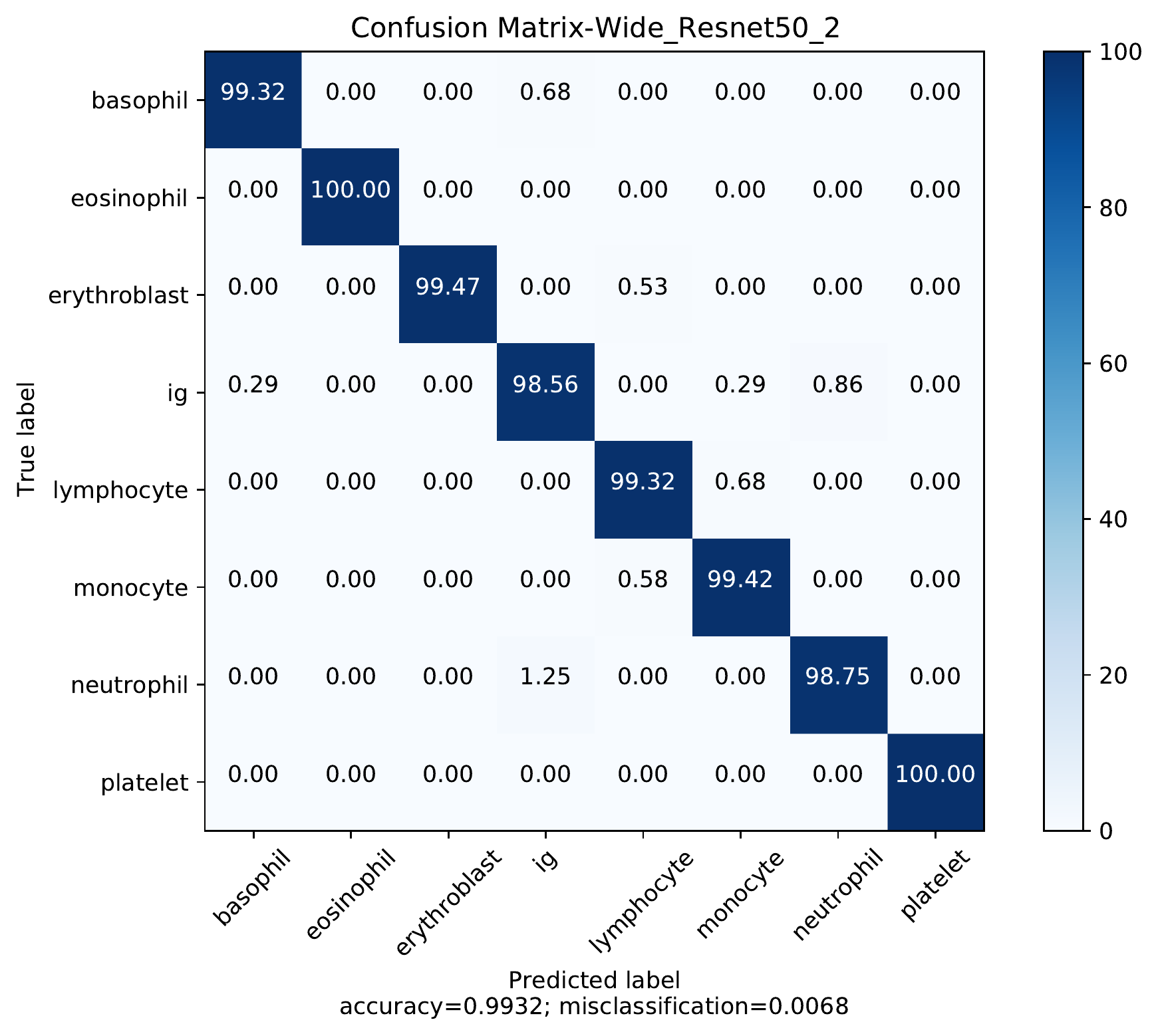}}
\end{center}
\vskip -0.3in
}{%
  \caption{Confusion Matrix for Wide ResNet-50\_2}%
  \label{conf_wide50}
}
\ffigbox{%
\centering
\begin{center}
\centerline{\includegraphics[width=0.95\columnwidth]{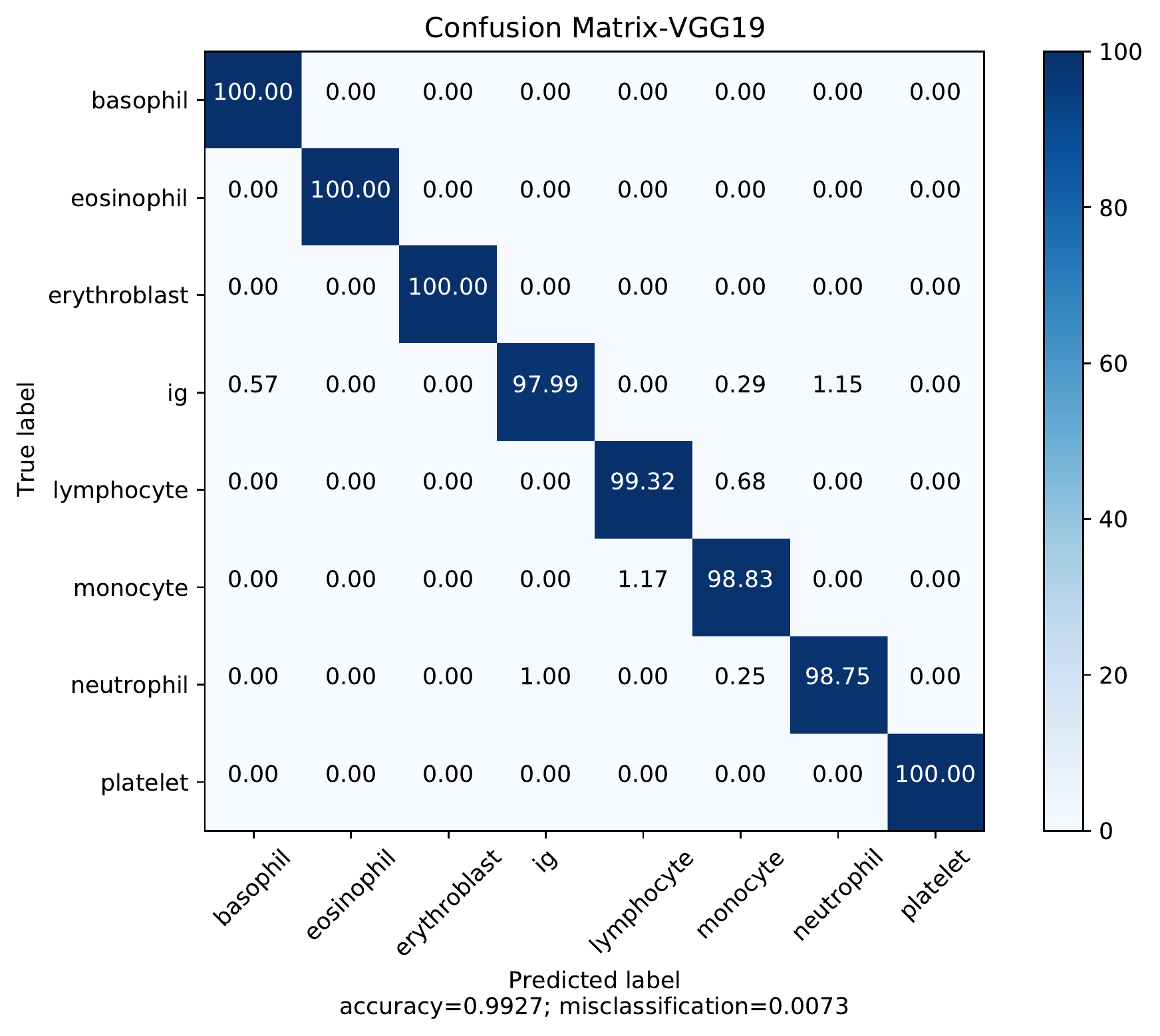}}
\end{center}
\vskip -0.3in
}{%
  \caption{Confusion Matrix for VGG-19}%
  \label{conf_vgg19}
}
\end{floatrow}
\end{figure}

\begin{figure}[h]
\begin{floatrow}
\ffigbox{%
\begin{center}
\centerline{\includegraphics[width=0.95\columnwidth]{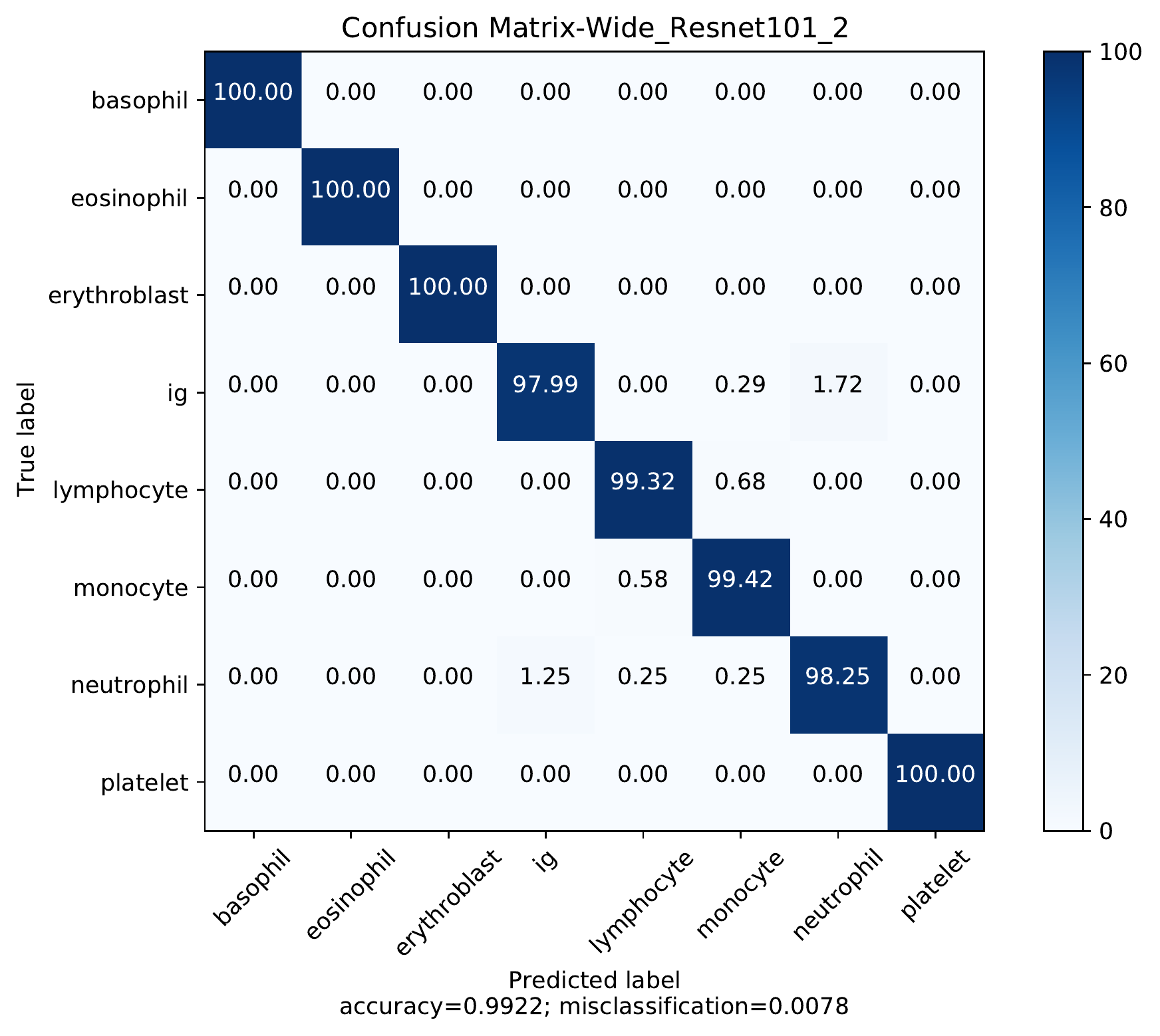}}
\end{center}
\vskip -0.3in
}{%
  \caption{Confusion Matrix for Wide ResNet-101\_2}%
  \label{conf_wide101}
}
\ffigbox{%
\centering
\begin{center}
\centerline{\includegraphics[width=0.95\columnwidth]{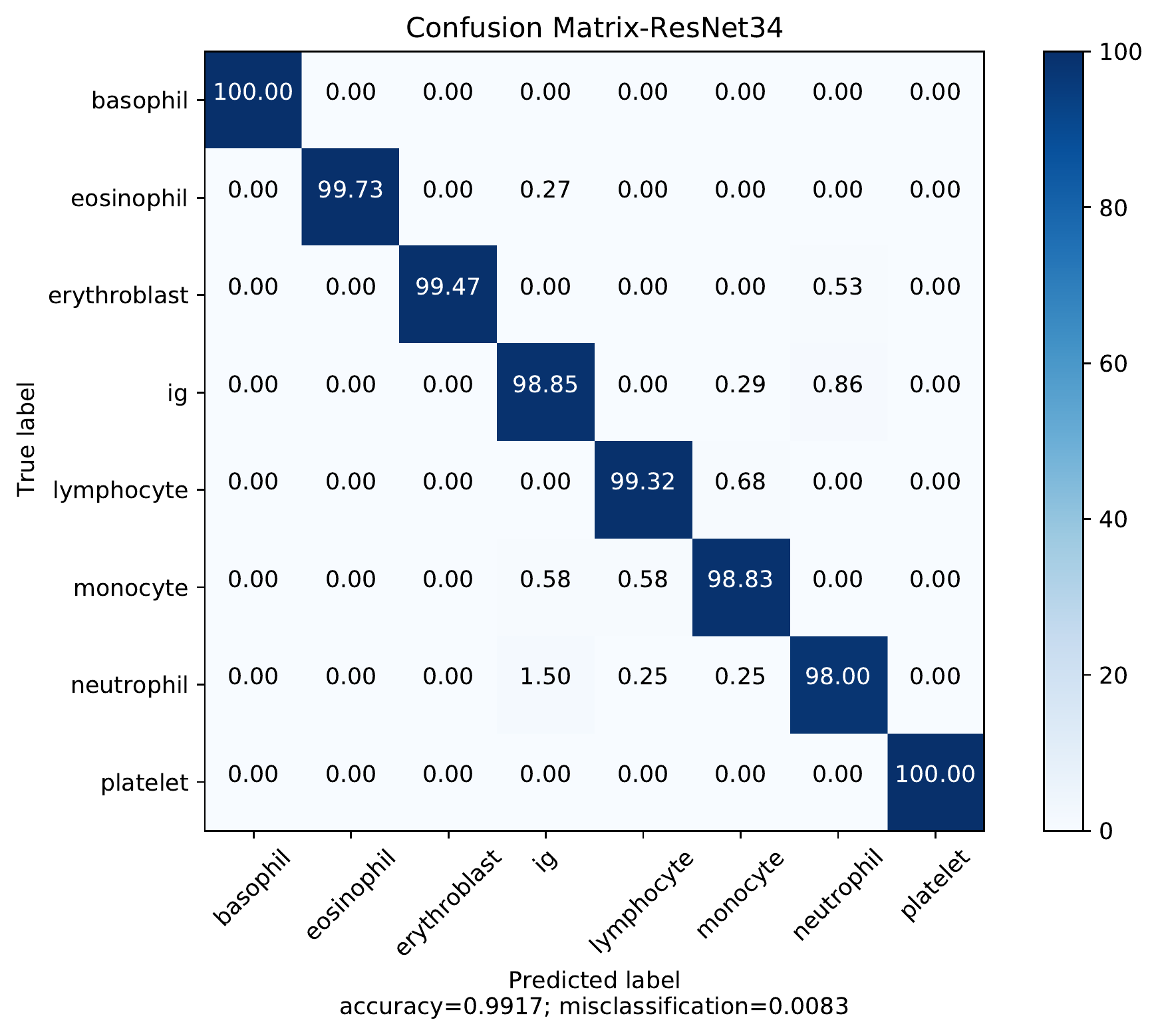}}
\end{center}
\vskip -0.3in
}{%
  \caption{Confusion Matrix for ResNet-34}%
  \label{conf_resnet34}
}
\end{floatrow}
\end{figure}

\begin{figure}[t]
\begin{floatrow}
\ffigbox{%
\begin{center}
\centerline{\includegraphics[width=\columnwidth]{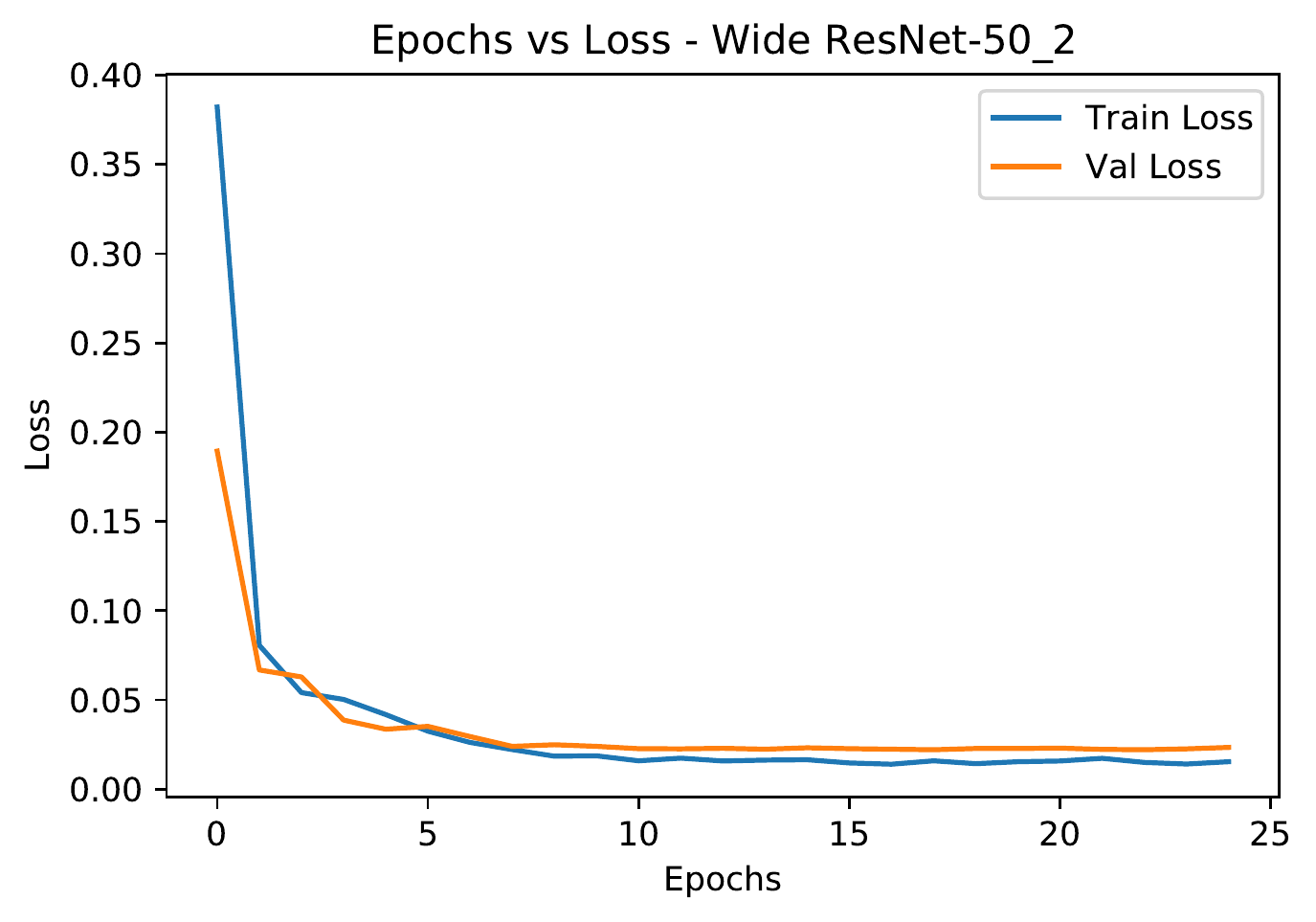}}
\end{center}
\vskip -0.4in
}{%
  \caption{Loss vs Epochs plot for Wide ResNet-50\_2}%
  \label{loss_wide50}
}
\ffigbox{%
\centering
\begin{center}
\centerline{\includegraphics[width=\columnwidth]{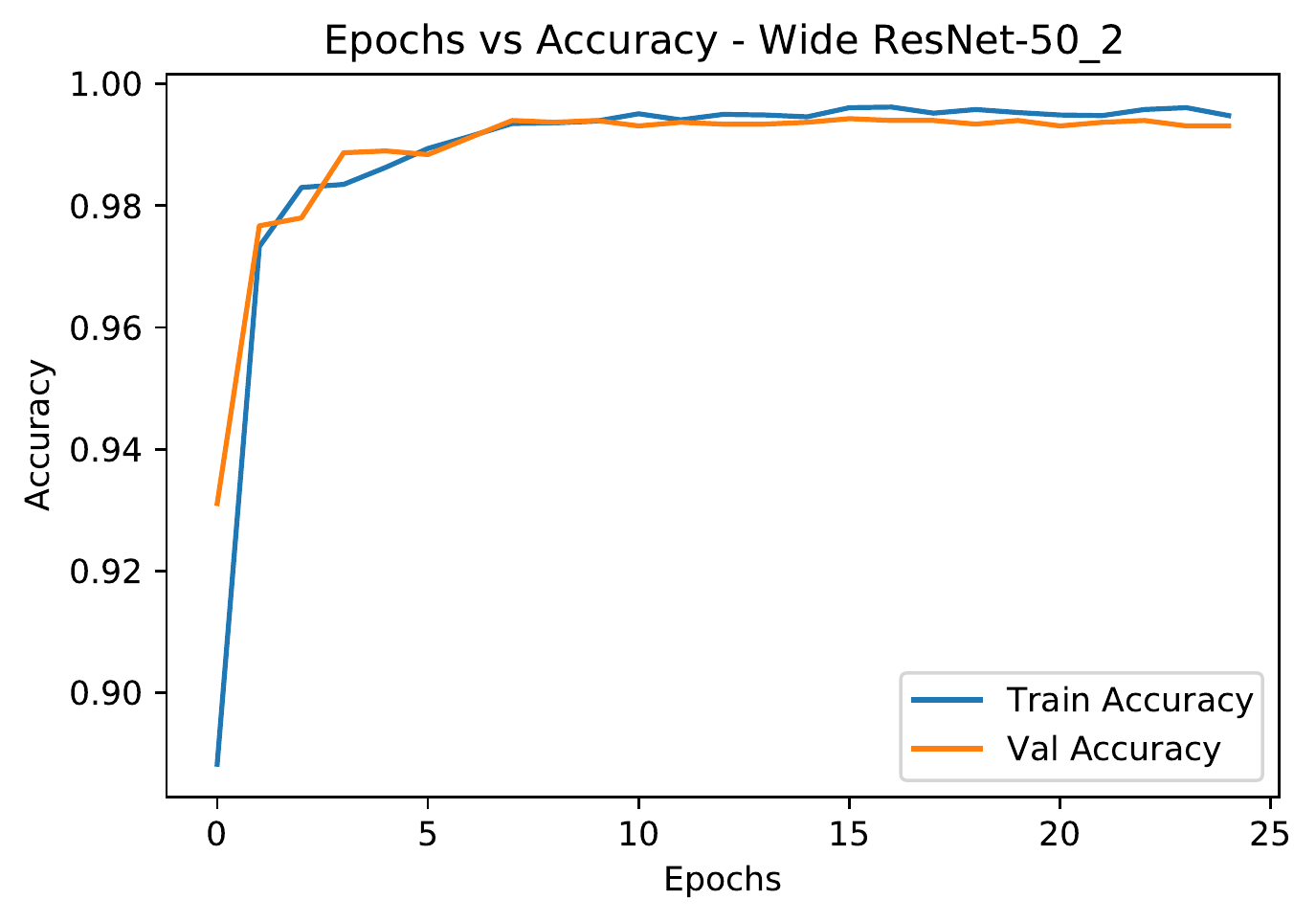}}
\end{center}
\vskip -0.4in
}{%
  \caption{Accuracy vs Epochs plot for Wide ResNet-50\_2}%
  \label{acc_wide50}
}
\end{floatrow}
\vskip -0.1in
\end{figure}

\begin{figure}
\begin{floatrow}
\ffigbox{%
\begin{center}
\centerline{\includegraphics[width=\columnwidth]{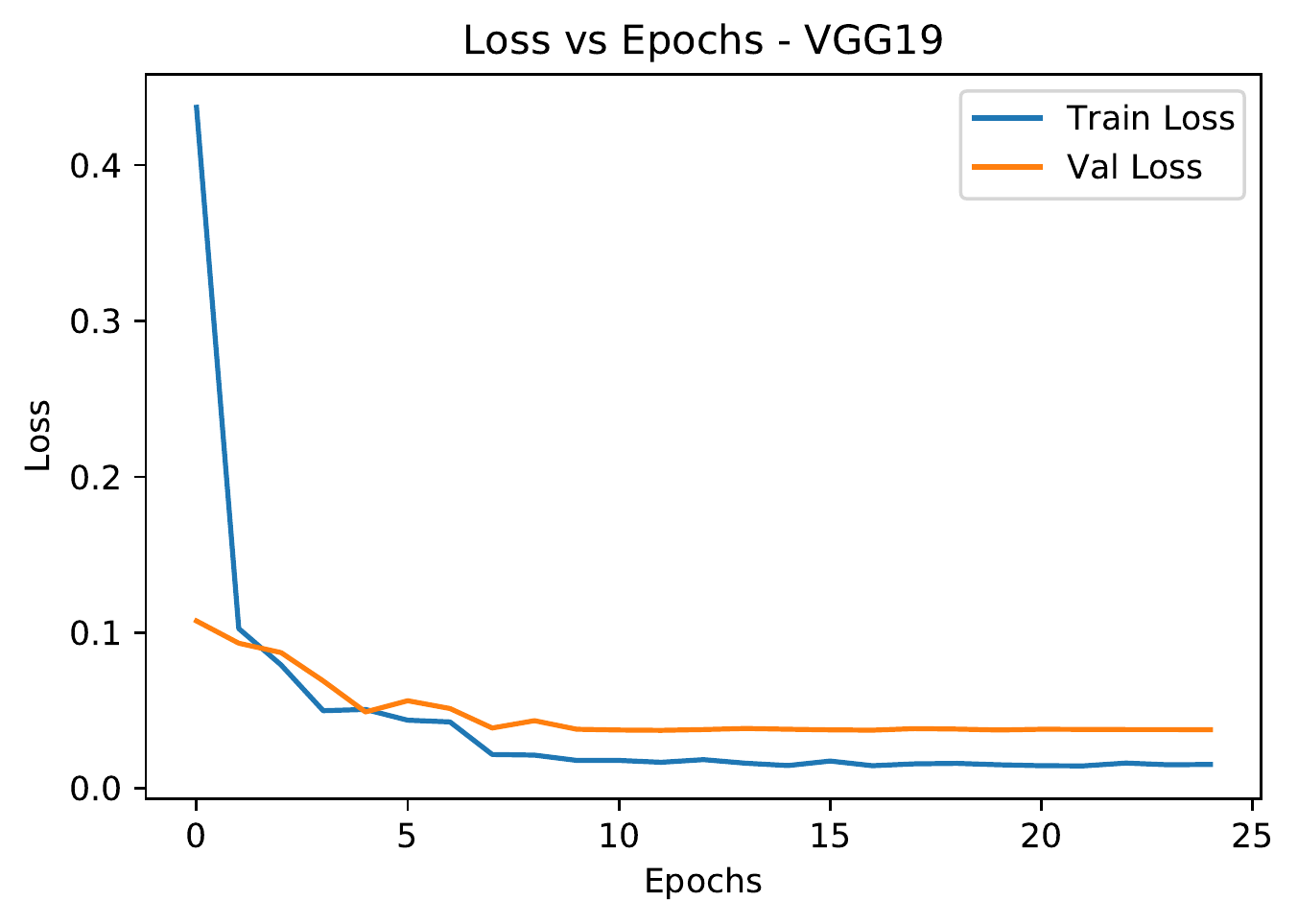}}
\end{center}
\vskip -0.3in
}{%
  \caption{Loss vs Epochs Plot for VGG-19}%
  \label{loss_vgg19}
}
\ffigbox{%
\centering
\begin{center}
\centerline{\includegraphics[width=\columnwidth]{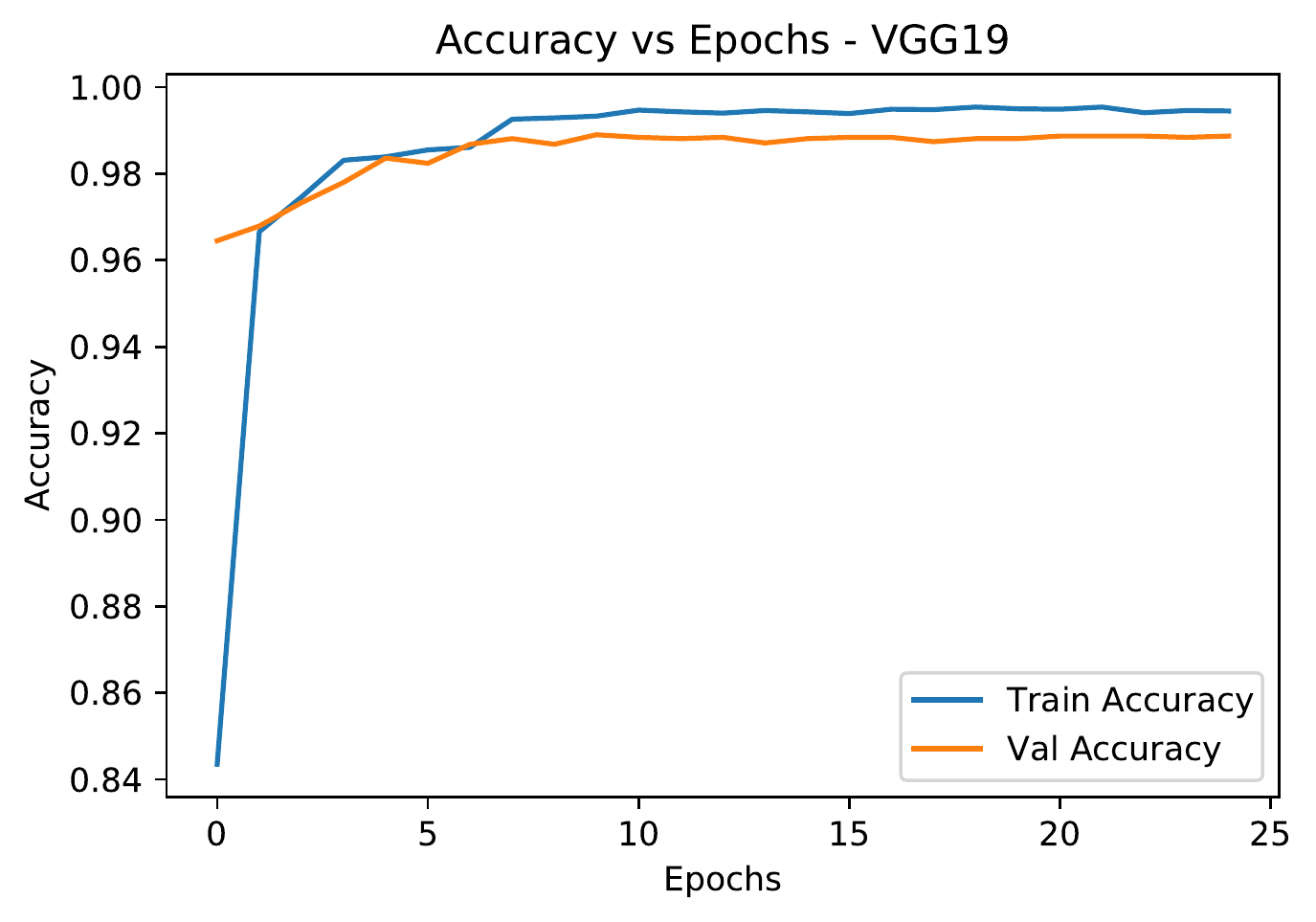}}
\end{center}
\vskip -0.3in
}{%
  \caption{Accuracy vs Epochs Plot for VGG-19}%
  \label{acc_vgg19}
}
\end{floatrow}
\end{figure}

\begin{figure}[t]
\begin{floatrow}
\ffigbox{%
\begin{center}
\centerline{\includegraphics[width=\columnwidth]{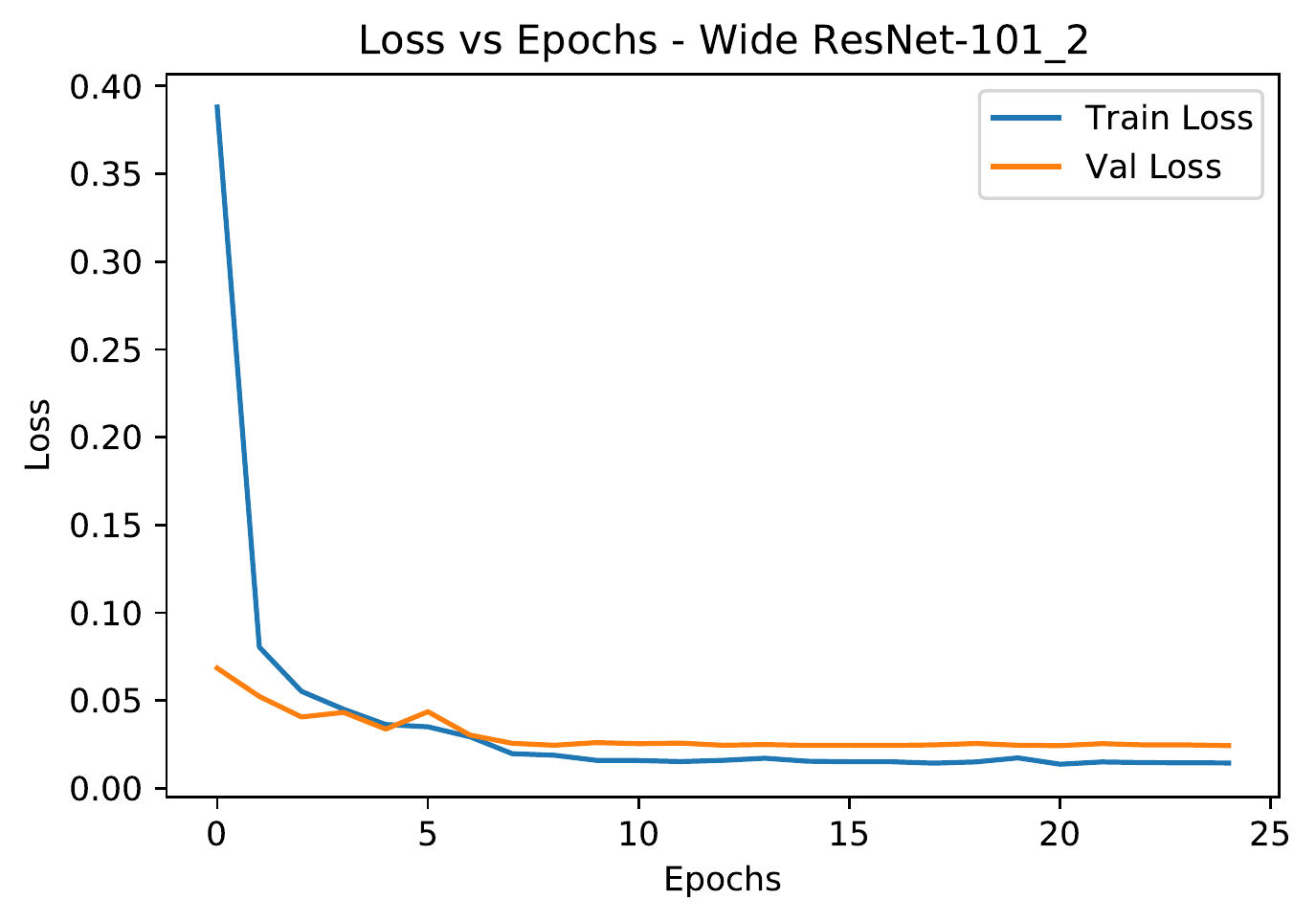}}
\end{center}
\vskip -0.4in
}{%
  \caption{Loss vs Epochs Plot for Wide ResNet-101\_2}%
  \label{loss_wide101}
}
\ffigbox{%
\centering
\begin{center}
\centerline{\includegraphics[width=\columnwidth]{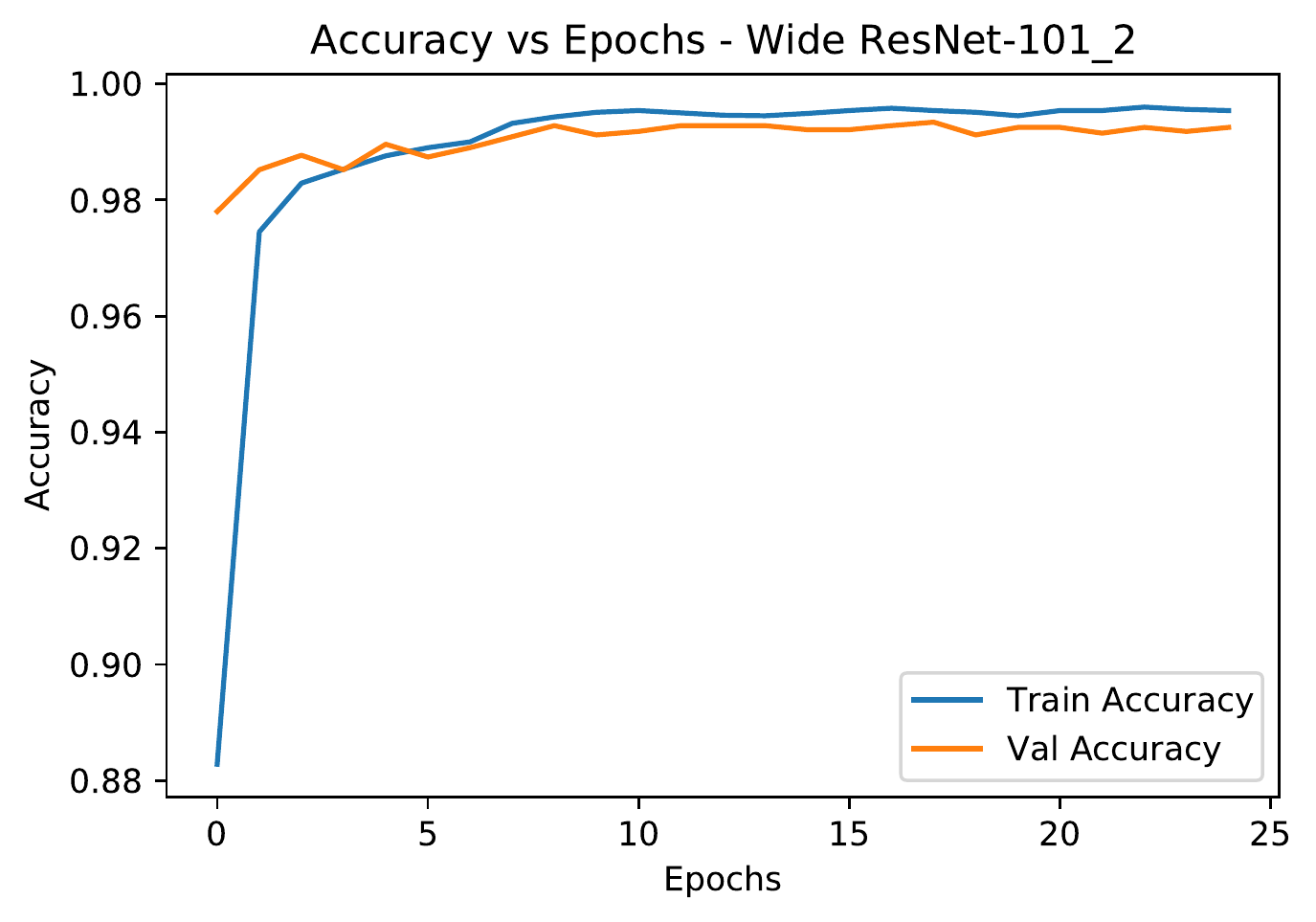}}
\end{center}
\vskip -0.4in
}{%
  \caption{Accuracy vs Epochs Plot for Wide ResNet-101\_2}%
  \label{acc_wide101}
}
\end{floatrow}
\vskip -0.1in
\end{figure}

\begin{figure}
\begin{floatrow}
\ffigbox{%
\begin{center}
\centerline{\includegraphics[width=\columnwidth]{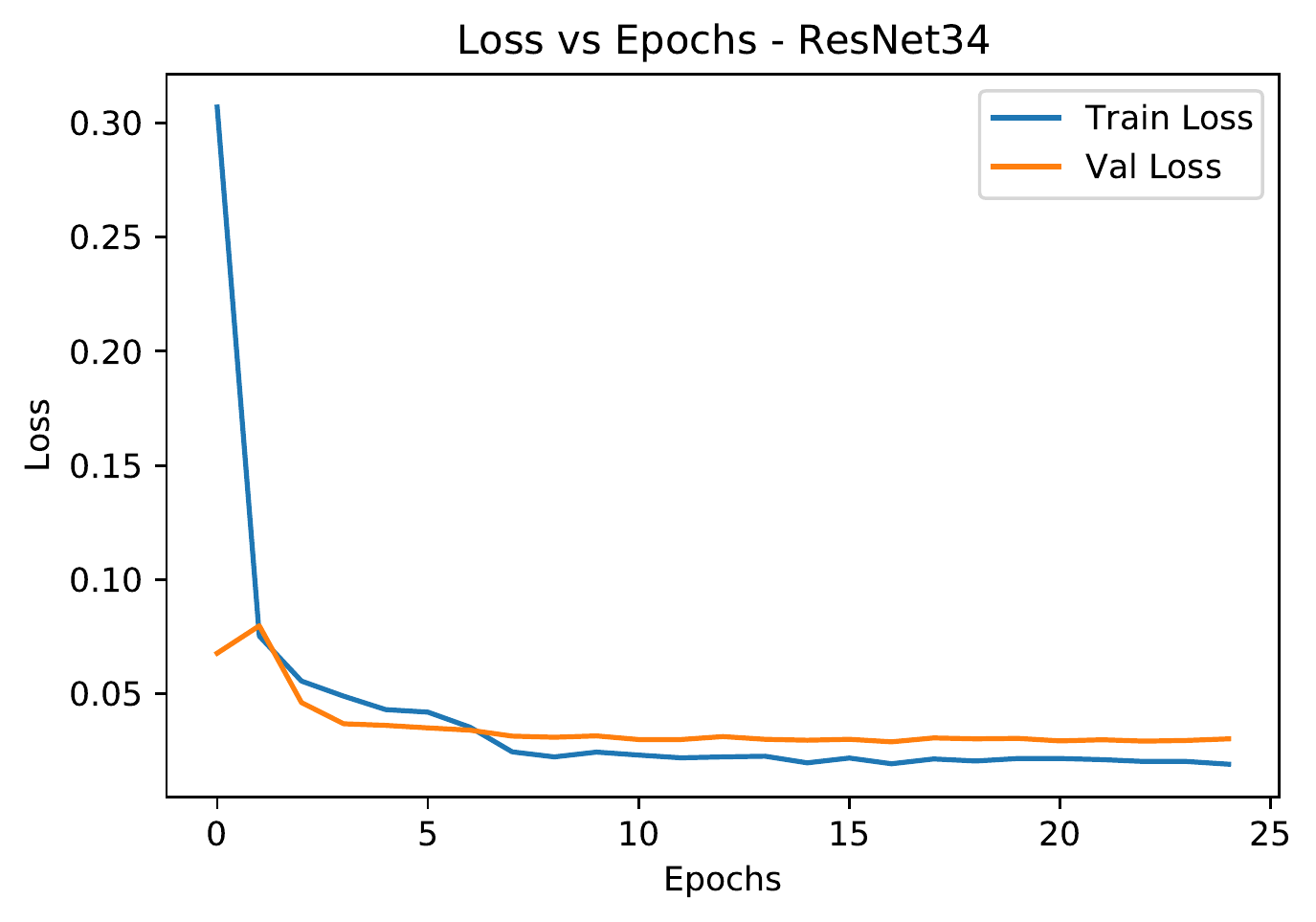}}
\end{center}
\vskip -0.3in
}{%
  \caption{Loss vs Epochs Plot for ResNet-34}%
  \label{loss_resnet}
}
\ffigbox{%
\centering
\begin{center}
\centerline{\includegraphics[width=\columnwidth]{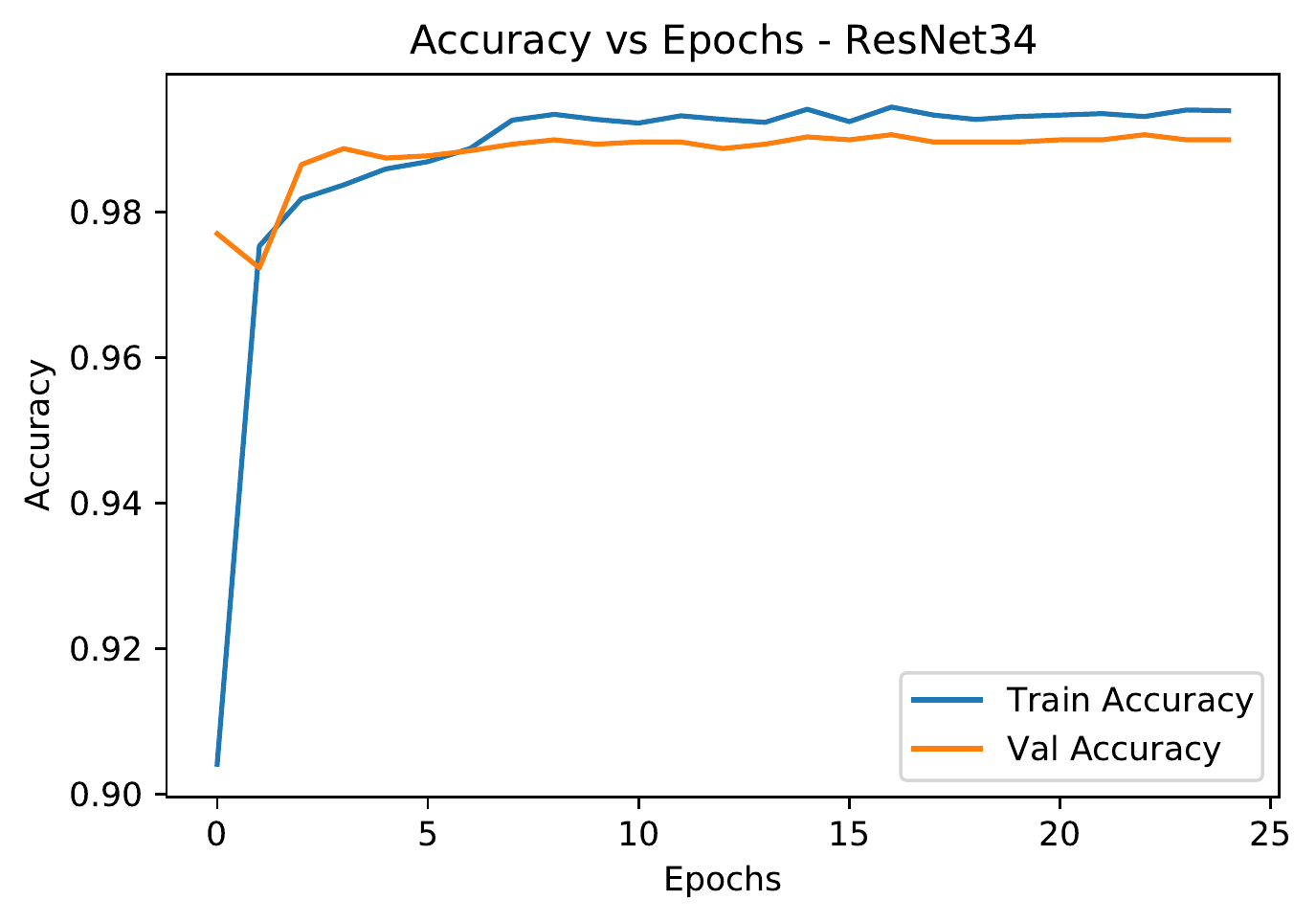}}
\end{center}
\vskip -0.3in
}{%
  \caption{Accuracy vs Epochs Plot for ResNet-34}%
  \label{acc_resnet34}
}
\end{floatrow}
\end{figure}

In this section, the training and validation plots for accuracy (Figures \ref{acc_wide50}, \ref{acc_vgg19}, \ref{acc_wide101}, \ref{acc_resnet34}), and loss (Figures \ref{loss_wide50}, \ref{loss_vgg19}, \ref{loss_wide101}, \ref{loss_resnet}) are presented along with the confusion matrices (Figures \ref{conf_wide50}, \ref{conf_vgg19}, \ref{conf_wide101}, \ref{conf_resnet34}) of our four top performing models are presented.

\end{document}